\documentclass[11pt]{article}
\usepackage[english]{babel}
\usepackage[letterpaper,top=2.5cm,bottom=2.5cm,left=3.5cm,right=3.5cm,marginparwidth=1.75cm]{geometry}

\usepackage{authblk}
\usepackage{microtype}
\usepackage{booktabs}
\usepackage[utf8]{inputenc}
\usepackage{multirow}
\usepackage{caption}
\usepackage{subcaption}
\usepackage{stmaryrd}


\usepackage{placeins}

\usepackage[ruled,vlined]{algorithm2e}

\SetAlgoNoLine
\DontPrintSemicolon

\usepackage{lmodern}
\usepackage{amsmath,amssymb,mathtools,amsthm,bm}

\usepackage{natbib}
\usepackage{enumitem}
\usepackage[dvipsnames]{xcolor}
\usepackage{graphicx}
\definecolor{navy}{RGB}{0,0,128}
\usepackage[colorlinks=true, allcolors=navy]{hyperref}
\usepackage[capitalize,noabbrev]{cleveref}
\usepackage[parfill]{parskip}

\usepackage[most]{tcolorbox}

\newtcolorbox{subtlebox}{
    colback=gray!8,          
    colframe=gray!8,         
    arc=0mm,                 
    boxrule=0pt,             
    top=4pt, bottom=4pt,   
    left=4pt, right=4pt,
    after skip=15pt,
    breakable                
}

\newtheorem{innerthm}{Theorem}[section]
\newtheorem{innerprop}[innerthm]{Proposition}
\newtheorem{innerlem}[innerthm]{Lemma}

\crefname{innerthm}{Theorem}{Theorems}
\crefname{innerprop}{Proposition}{Propositions}
\crefname{innerlem}{Lemma}{Lemmas}

\newenvironment{theorem}[1][]
  {\begin{subtlebox}\begin{innerthm}[#1]}
  {\end{innerthm}\end{subtlebox}}

\newenvironment{proposition}[1][]
  {\begin{subtlebox}\begin{innerprop}[#1]}
  {\end{innerprop}\end{subtlebox}}

\newenvironment{lemma}[1][]
  {\begin{subtlebox}\begin{innerlem}[#1]}
  {\end{innerlem}\end{subtlebox}}

\usepackage{pifont}

\usepackage[font=small]{caption}
\usepackage{subcaption}

\usepackage{etoolbox}
\AtBeginEnvironment{table}{\small}

\newcommand{\appendixtitle}{%
  \par\noindent{\LARGE\bfseries Appendix\par}%
}

\usepackage{titling}
\date{}

\pretitle{\vspace{-0.5cm}\centering \rule{\linewidth}{2pt}\\[0.3ex]\LARGE\bfseries}
\posttitle{\\[0ex]\rule{\linewidth}{1pt}\vspace{2ex}}

\newcommand{\mytitle}{Martingale Posterior Cox Processes}
\hypersetup{
    pdftitle={\mytitle},
}
\title{\mytitle}

\DeclareRobustCommand{\authorblock}[2]{%
  \begin{minipage}[t]{0.4\linewidth}
    \centering
    {\large\bfseries #1\par}
    {\normalfont\large #2\par}
  \end{minipage}%
}

\author{%
  \authorblock{Joshua Rooijakkers}{Imperial College London}%
  \hspace{1em}%
  \authorblock{Edwin Fong}{University of Hong Kong}%
}

\begin{document}
\maketitle

\begin{abstract}
    \noindent Cox processes provide a principled Bayesian nonparametric framework for modeling event-based data through an underlying intensity function. However, existing inference methods for estimating the intensity typically rely on computationally expensive Markov chain Monte Carlo sampling or approximate inference techniques. In addition, to ensure tractability, these methods often define priors over indirect latent structures, making them challenging to interpret. In this paper, we introduce the Martingale Posterior Cox Process (MPCP), which recasts Bayesian intensity estimation as inference over unobserved events given the observed event history. This perspective avoids the need to define complex priors while retaining the flexibility of nonparametric modeling. We propose a scalable, GPU-friendly copula-based predictive update, and provide strong theoretical guarantees for the convergence of the predictive resampling algorithm. Empirically, we show that MPCP provides excellent performance in both simulations and real data at a fraction of the computational cost of competing Bayesian approaches.
\end{abstract}

\section{Introduction}

Poisson processes are a fundamental tool for modeling event-based data across a wide range of fields, including medicine, economics, engineering and biology. A central objective in these applications is to estimate the potentially inhomogeneous intensity function governing the occurrence of events. When the intensity function is itself unknown, it is commonly modeled as a stochastic process and inferred using nonparametric Bayesian methods, giving rise to \textit{Cox processes} \citep{cox1955some}. Their flexibility has made Cox processes a standard modeling framework in applications ranging from epidemiology \citep{benevs2003application} and forestry \citep{heikkinen1999modeling} to finance \citep{basu2002cox} and neuroscience \citep{cunningham2007inferring}.

Despite their widespread use, scalable and accurate inference for Cox process intensity estimation remains challenging. Existing approaches typically use a Gaussian process (GP) prior with a positive link function or rely on Dirichlet process mixture models (DPMMs). These methods require either computationally expensive Markov chain Monte Carlo (MCMC) sampling \citep{moller1998log, adams2009tractable, kottas2007bayesian} or approximate inference \citep{gunter2014efficient, lloyd2015variational}. Additionally, their prior structures are difficult to interpret: they are defined either in a latent space transformed by a nonlinear link function (for GPs) or via infinite-dimensional mixtures (for DPMMs), primarily introduced for computational convenience. As a result, these limitations hinder the applicability of Cox process models in settings where scalability and interpretability are essential.

In this paper, we introduce the \textit{Martingale Posterior Cox Process} (MPCP), a scalable alternative that avoids specifying priors on latent spaces. Instead, we need only elicit a prior directly on mean event counts. We then recast Bayesian intensity estimation by characterizing its posterior uncertainty through unobserved event configurations given the observed events via the framework of martingale posterior distributions. Crucially, this framework allows for a GPU-friendly and parallelizable algorithm with closed-form updates, eliminating the need for MCMC or approximate inference techniques. Our main contributions are as follows:
\begin{enumerate}
    \item We propose a class of copula-based predictive updates tailored to Cox process intensity estimation within the martingale posterior framework.
    \item We provide convergence guarantees for an MCMC-free martingale posterior algorithm for uncertainty in the intensity function. 
    \item We show empirically that MPCP provides excellent predictive performance and reliable uncertainty quantification on both simulated and real data, while requiring substantially less computational time than both MCMC and variational inference approaches.
\end{enumerate}

\section{Related Work}

Nonparametric Bayesian inference for point processes has been studied extensively in the literature. The dominant approach models the prior on the intensity function through a latent GP \citep{williams2006gaussian} transformed by a positive link function to ensure non-negativity. Common choices include the exponential \citep{moller1998log}, sigmoidal \citep{adams2009tractable}, and square \citep{mccullagh2006permanental} link functions. Since the Poisson likelihood is generally non-conjugate to the GP prior, exact Bayesian inference is typically intractable. Consequently, inference relies on MCMC methods \citep{moller1998log, adams2009tractable, gunter2014efficient, samo2015scalable, john2018large}, Laplace approximations \citep{walder2017fast, kim2021fast, sellier2023sparse, sun2024nonstationary}, or variational inference techniques \citep{lloyd2015variational, lian2015multitask, donner2018efficient, aglietti2019structured}. While these methods provide flexible intensity estimates, they require inference over latent functions whose relationship to the observed point pattern is indirect, and scalability remains a persistent challenge for large datasets.

An alternative Bayesian nonparametric framework models the intensity through DPMMs \citep{escobar1995bayesian}. In particular, \citet{kottas2006dirichlet, kottas2007bayesian} consider a mixture of Beta kernel functions whose mixing distribution is assigned a Dirichlet process prior. However, posterior inference still relies on computationally intensive MCMC, and the induced prior is specified indirectly through an infinite-dimensional mixing distribution, which can be difficult to interpret. Our approach is closely related to DPMMs, but instead directly defines a predictive distribution over future event configurations, avoiding latent-function priors as well as MCMC-based or other approximate inference schemes.

\section{Background} \label{sec:background}

Our method applies the framework of martingale posterior distributions to Cox processes. In this section, we briefly explain the concepts relevant to understanding this interface.

\subsection{Cox Processes}

A Cox process, or a doubly stochastic Poisson process, is a Poisson process whose intensity function is itself random. Specifically, let $\lambda : \mathcal{X} \rightarrow \mathbb{R}^{+}$ denote a stochastic intensity function defined on a domain $\mathcal{X} \subset \mathbb{R}^{d}$ where $\int_{\mathcal{X}}\lambda(x) \textup{d}x < \infty$ a.s. Conditional on $\lambda$, the process reduces to an inhomogeneous Poisson process: for any measurable subset $\mathcal{A} \subseteq \mathcal{X}$, the number of events in $\mathcal{A}$ follows a Poisson distribution with mean $\int_{\mathcal{A}} \lambda(x) \,\textup{d}x$, and event counts in disjoint subsets are independent. Let $x_{1:n} = \{x_i\}_{i=1}^{n}$ denote the unordered set of $n$ observed events, where $x_i \in \mathcal{X}$. Conditional on the intensity function $\lambda$, the likelihood is given by
\begin{align}
    f(x_{1:n} \mid  \lambda) = \exp\left\{-\int_{\mathcal{X}} \lambda(x) \textup{d}x\right\} \prod_{i=1}^n \lambda(x_i). \label{eq:likelihood_lambda}
\end{align}
Given a prior probability distribution $\Pi$ over intensity functions, Bayes' theorem yields the posterior
\begin{align}
    \Pi(\textup{d}\lambda \mid  x_{1:n}) \propto \Pi(\textup{d}\lambda) \exp\left\{-\int_{\mathcal{X}} \lambda(x) \textup{d}x\right\}\prod_{i=1}^n \lambda(x_i)\label{eq:posterior_lambda}
\end{align}
This posterior is sometimes referred to as \textit{doubly intractable}, due to the potentially intractable integral in the likelihood function \citep{murray2007advances}. Posterior inference then traditionally relies on MCMC methods or approximate inference techniques.

\subsection{Martingale Posterior Distributions}

Rather than constructing the posterior distribution directly from a prior and likelihood, we instead turn to martingale posteriors, which enable inference through a sequence of predictive distributions \citep{fong2023martingale}. We first introduce the martingale posterior for a general parameter of interest $\theta$ and later show how it can be used to estimate the intensity rate $\lambda$ in \Cref{sec:methodology}. In the classical Bayesian framework, the predictive distribution for a future observation $X_{n+1}$ given observed data $x_{1:n}$ is
\begin{align}
    p(x_{n+1}\mid x_{1:n}) = \int f(x_{n+1} \mid \theta) \, \pi(\theta \mid x_{1:n}) \textrm{d}\theta, \label{eq:predictive}
\end{align}
where $\pi(\theta \mid x_{1:n}) \propto \pi(\theta) \prod_{i=1}^nf(x_i \mid \theta)$ denotes the posterior distribution induced by likelihood $f(x\mid\theta)$ and prior $\pi(\theta)$. The martingale posterior framework instead recasts Bayesian inference as an imputation problem. Rather than specifying a prior and likelihood, one directly elicits predictive distributions for the unobserved remainder of the population. In the i.i.d. setting, this corresponds to the future observations $X_{n+1:\infty}$, conditional on the observed sample $x_{1:n}$. The resulting joint predictive distribution can be factorized as
\begin{align*}
    p(x_{n+1:N} \mid x_{1:n}) = \prod_{i=n+1}^N p_{i-1}(x_i),
\end{align*}
where $p_i$ denotes the one-step-ahead predictive density with cumulative distribution function (CDF) $P_i$. Given the completed population $(x_{1:n}, X_{n+1:N})$, the quantity of interest can be written as $\theta(x_{1:n},X_{n+1:N})$. Uncertainty in the imputed observations $X_{n+1:N}$ then induces a distribution over $\theta$, which defines the martingale posterior. Further details are provided in Appendix~\ref{app:notation_preliminaries}.

Given a likelihood $f(x\mid\theta)$ and prior $\pi(\theta)$, the predictive is defined as in \eqref{eq:predictive}. More generally, the martingale posterior distribution considers sequences of predictive distributions beyond \eqref{eq:predictive}, and thus generalizes standard Bayesian inference. Given a sequence of predictive CDFs $P_n(x), P_{n+1}(x), \dots$, the unobserved population can be imputed via a sequential scheme termed predictive resampling \citep{Fortini2020, fong2023martingale}:
\begin{align*}
    X_{n+1} \sim P_n(x), \quad X_{n+2} \sim P_{n+1}(x), \quad \dots .
\end{align*}
Since inference over an infinite population is infeasible in practice, predictive resampling is truncated at some large $N>n$. A sufficient condition for the existence of the martingale posterior is that the induced sequence is conditionally identically distributed (c.i.d.; \citealp{berti2004limit}). Specifically, the sequence $X_{n+1}, X_{n+2}, \dots$ is c.i.d. if
\begin{align*}
    P(X_{i+k} \leq x \mid x_{1:i}) = P_i(x), \qquad \forall k > 0.
\end{align*}
Under this condition, the predictive distributions form a martingale and predictive resampling yields a well-defined posterior distribution.

\subsection{Bivariate Copula Updates} \label{sec:bivariate_copula_updates}

We now consider a concrete example of a c.i.d. sequence of predictive densities that is computationally tractable and does not require specification of a likelihood or prior distribution. Specifically, we consider the class of predictive densities based on bivariate copula densities introduced by \citet{hahn2018recursive}. A bivariate copula density is a bivariate probability density function (PDF) with uniform marginals, i.e. a function $c: [0,1]^2 \rightarrow \mathbb{R}^+$ satisfying $\int c(u,v) \textup{d}u = \int c(u, v) \textup{d}v = 1$. 
In the nonparametric setting, \citet{hahn2018recursive, fong2023martingale} propose a tractable sequence of copula-based updates inspired by the DPMM \citep{ferguson1973bayesian, escobar1995bayesian}. In general, a DPMM takes the form
\begin{align*}
    f(x \mid G) = \int K(x\mid\theta) \,G(\textup{d}\theta), \qquad G \sim \textup{DP}(c, G_0),
\end{align*}
where $\theta \in \Theta$ is a parameter vector, $G$ is assigned a Dirichlet process (DP) prior with base measure $G_0$ and concentration parameter $c > 0$, and $K(x\mid\theta)$ is a user-specified kernel. For example, \citet{kottas2006dirichlet, kottas2007bayesian} use beta kernels in their applications to Cox processes. By leveraging this framework, \citet{hahn2018recursive} derive a recursive update corresponding to the DPMM. In the univariate setting, the update is given by:
\begin{equation}
\begin{aligned}
    p_i(x) &= \left[1 - \alpha_i + \alpha_i c_{\phi}\{P_{i-1}(x), P_{i-1}(x_i)\}\right] p_{i-1}(x) \\
    P_i(x) &= (1-\alpha_i) P_{i-1}(x) + \alpha_i H_\phi\{P_{i-1}(x), P_{i-1}(x_i)\}, \label{eq:copula_updates}
\end{aligned}
\end{equation}
where $\alpha_i = (2-i^{-1})/(i+1)$, $c_{\phi}$ is a bivariate copula density with parameter $\phi$, and $H_\phi$ is its corresponding $h$-function, defined as $H_\phi(u,v) = \partial C_\phi(u,v) / \partial v$, where $C_\phi$ is the copula CDF. Notably, \citet{hahn2018recursive, fong2022predictive, fong2023martingale} show that the above update coincides exactly with the DPMM update from $p_0$ to $p_1$, but is different for $p_i$ with $i > 1$. Unlike the DPMM, however, the above updates are available in closed form, and we can thus calculate the posterior predictive density without MCMC. In the multivariate case, \citet{fong2023martingale} derive a similar update:
\begin{align}
    p_i(x) = \Bigl\{1 - \alpha_i + \alpha_i \prod_{j=1}^d c_{\phi_j} (u_{i-1}^{j}, v_{i-1}^{j}) \Bigr\} p_{i-1}(x), \label{eq:multivariate_copula_updates}
\end{align}
where $u_i^{j} = P_i(x^{j} \mid x^{1:j-1})$ and $v_i^j = P_i(x_{i+1}^j \mid x_{i+1}^{1:j-1})$, and $x^{j}$ and $x^{1:j}$ denote the $j$-th covariate and first $j$ covariates of $x$, respectively. This formulation requires a specific ordering of the covariate dimensions, with each dimension $j$ associated with its own copula $c_{\phi_j}$ parametrized by $\phi_j$. The updates for the conditional CDFs $u_i^j$ and $v_i^j$ are similarly tractable; details are provided in Appendix~\ref{app:notation_preliminaries}.

\section{Methodology} \label{sec:methodology}

Let $\gamma := \int_{\mathcal{X}} \lambda(x) \textrm{d}x$ and define the normalized intensity $s(x) := \lambda(x) / \gamma$, which is a PDF. We may then rewrite the likelihood in \eqref{eq:likelihood_lambda} as
\begin{align*}
    f(x_{1:n} \mid s, \gamma) = e^{-\gamma} \gamma^n \prod_{i=1}^n s(x_i).
\end{align*}
Since $(s, \gamma)$ provides an equivalent parametrization of $\lambda$, we can express the posterior in \eqref{eq:posterior_lambda} in terms of these quantities. Now suppose we place a factorized prior $\Pi(\textup{d}s,\textup{d}\gamma) = \Pi(ds)\,\pi(\gamma)\textup{d}\gamma$ where $\pi(\gamma)$ is a PDF on $(0,\infty)$. The posterior then factorizes as
\begin{align}
    \Pi(\textup{d}s, \textup{d}\gamma \mid x_{1:n}) &= \Pi(\textup{d}s \mid x_{1:n}) \,\pi(\gamma\mid n)\textup{d}\gamma, \label{eq:factorized_posteriors}
\end{align}
where $\Pi(\textup{d}s\mid x_{1:n}) \propto \Pi(\textup{d}s)\prod_{i=1}^n s(x_i)$ and $\pi(\gamma\mid n) \propto \pi(\gamma)\,e^{-\gamma}\gamma^n$. We use this equivalent approach for predictive resampling in the MPCP model.

\subsection{Predictive Resampling in Cox Processes} \label{sec:predictive_resampling_cox}

The factorized posteriors in \eqref{eq:factorized_posteriors} admit an interesting predictive interpretation. The term $\Pi(ds\mid x_{1:n})$ is the posterior over the normalized intensity, and thus captures how the data inform the spatial structure of the intensity. In contrast, $\pi(\gamma\mid n)$ depends only on the total number of observations $n$ and therefore governs the overall scale of the process, effectively estimating the expected total count.

\begin{figure}[t]
  \centering
  \includegraphics[width=0.7\textwidth]{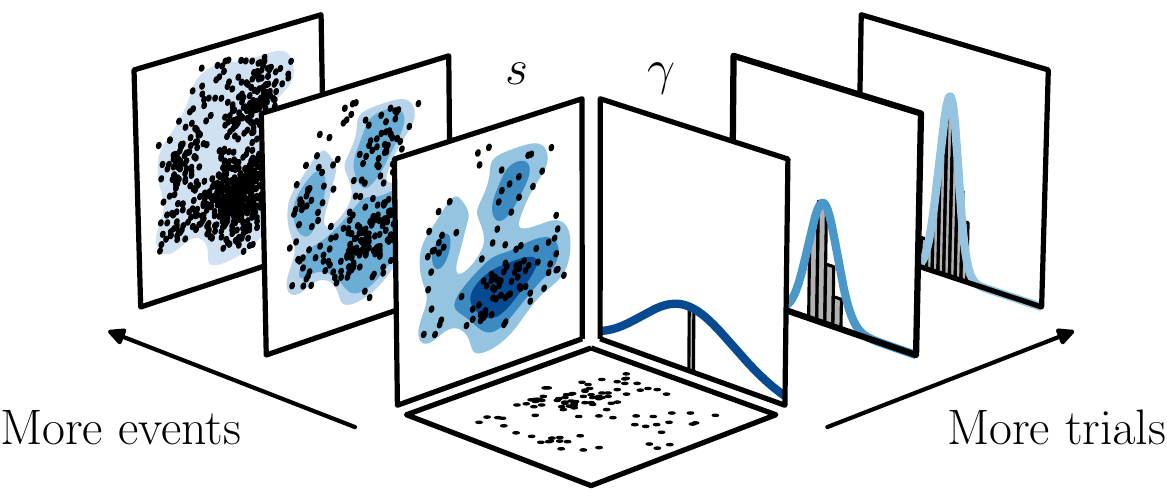}
  \caption{Predictive view of Cox processes, where darker shades of blue indicate greater uncertainty. The posterior intensity given the observed data (bottom) decomposes into a density (left) and count (right) component. While uncertainty in the density decreases with the number of observed events, uncertainty in the count decreases with the number of repeated realizations of the process.}
  \label{fig:predictive_view}
\end{figure}

From a predictive perspective, these two sources of uncertainty are reduced in orthogonal ways (see \Cref{fig:predictive_view}). Uncertainty about the density $s$ decreases as more event locations are observed within a realization of the process, and predictive resampling can therefore be achieved by imputing additional event locations. For the spatial setting, this can be viewed as `waiting' after observing $x_{1:n}$ so that more events occur within the spatial region. In contrast, uncertainty about the average count $\gamma$ is reduced only through observing repeated realizations of the whole process, each providing information about the total count. The overall uncertainty in the intensity function $\lambda = \gamma s$ is induced by these two orthogonal sources of uncertainty. 

Predictive resampling of $\gamma$ would require imputing counts across multiple independent realizations, which is typically infeasible in practice since only a single realization of the process is observed. Fortunately, a simple and elegant solution is to place a conjugate prior on $\gamma$. Specifically, if $\pi(\gamma) = \Gamma(\gamma ; \alpha, \beta)$, where $\Gamma(\,\cdot\,;\alpha, \beta)$ is the gamma density with shape parameter $\alpha$ and rate parameter $\beta$, then the posterior is given by $\pi(\gamma \mid n) = \Gamma(\gamma ; \alpha + n, \beta + 1)$. This provides a simple and interpretable mechanism for encoding prior beliefs about the expected total count. When such prior information is unavailable, one may instead adopt the noninformative prior, $\pi(\gamma) \propto {\gamma^{-1}}$, which is also conjugate and yields the posterior $\pi(\gamma \mid n) = \Gamma(\gamma ; n, 1)$.

For the density $s$, we can perform predictive resampling by leveraging the nonparametric copula update introduced in \Cref{sec:bivariate_copula_updates}. Here, we will be taking the value $N \to \infty$ within the observed realization of the process, where uncertainty in $s$ is driven by the values $X_{n+1:\infty}$. In the next section, we will rely on the martingale posterior theory to show convergence under predictive resampling, where we tailor the results for the Cox process setting.  

\subsection{Copula Updates on Bounded Domains}

\begin{figure}[t]
  \centering
  \includegraphics[width=0.7\textwidth]{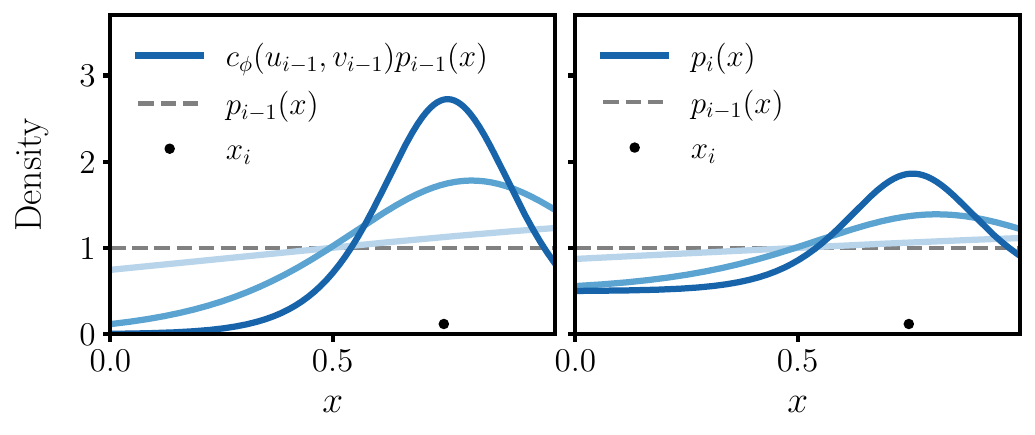}
  \caption{{Frank copula predictive update.} Left: copula kernel $k_i(x;x_i) = c_\phi(u_{i-1}, v_{i-1})\,p_{i-1}(x)$ with $u_{i-1} = P_{i-1}(x)$ and $v_{i-1} = P_{i-1}(x_i)$. Right: updated predictive $p_i(x)$ for $x_i = 0.75$, $\alpha_i = 0.5$ and $\phi = 1, 5, 10$ (\textcolor[RGB]{198, 219, 239}{\rule[0.5ex]{0.5cm}{1.5pt}},\textcolor[RGB]{107, 174, 214}{\rule[0.5ex]{0.5cm}{1.5pt}},\textcolor[RGB]{33, 113, 181}{\rule[0.5ex]{0.5cm}{1.5pt}}).}
  \label{fig:frank_copula}
    \vspace*{-0.1cm}
\end{figure}

For the density component, the bivariate copula updates described in \Cref{sec:bivariate_copula_updates} provide a flexible, nonparametric way for posterior learning. The choice of copula is particularly important in our setting, where the domain is often bounded and multivariate. Indeed, intensity estimation near domain boundaries is a well-known challenge in the literature, referred to as the \emph{boundary bias} problem \citep{diggle2013statistical}. 
Although previous work has successfully employed the Gaussian copula \citep{hahn2018recursive, fong2023martingale}, this choice is not suitable for bounded domains, as its evaluation vanishes at the boundaries. We therefore adopt the Frank copula, which does not suffer from this limitation and is available in closed form. Its density and corresponding $h$-function are given by
\begin{align*}
    c_\phi(u, v) &= \frac{-\phi (e^{-\phi} - 1) e^{-\phi(u+v)}}{(e^{-\phi} - 1 + (e^{-\phi u} - 1)(e^{-\phi v} - 1))^2},\\
    H_\phi(u,v) &= \frac{e^{-\phi v}(e^{-\phi u}-1)}{e^{-\phi}-1 + (e^{-\phi u} - 1)(e^{-\phi v} - 1)}\cdot
\end{align*}
Here, $\phi > 0$ controls the degree of localization of the update, where larger values yield a stronger peak. The copula update can thus be interpreted as an online kernel density estimator (KDE) adapted to bounded domains. This is illustrated in \Cref{fig:frank_copula} for $d=1$, where we plot both the copula kernel $k_i(x; x_i):= c_\phi(P_{i-1}(x), P_{i-1}(x_i))\, p_{i-1}(x)$ and the resulting updated density for increasing values of $\phi$. We see that $k_i(x; x_i)$ appears as a kernel with a peak near the observed value $x_i$. However, in contrast to standard KDEs, the copula updates are data-dependent and form a c.i.d. sequence, enabling predictive uncertainty quantification.

To begin, we formalize the intuition that the Frank copula is less susceptible to the boundary bias problem than the Gaussian copula. 

\begin{proposition}\label{prop:cop_limits}
For any $v \in (0,1)$,  the Gaussian copula density $c_\rho(u,v)$ with $\rho \in (0,1)$ satisfies
\begin{align*}
    \lim_{u \to 0} c_\rho(u,v) = 0, \quad \lim_{u \to 1} c_\rho(u,v) = 0.
\end{align*}
In contrast, the Frank copula for $\phi > 0$ satisfies
\begin{align*}
    \lim_{u \to 0} c_\phi(u,v) &= \frac{\phi e^{-\phi v}}{1-e^{-\phi}} > 0, \\
    \lim_{u \to 1} c_\phi(u,v) &= \frac{\phi e^{-\phi(1-v)}}{1- e^{-\phi}} > 0.
\end{align*}
\end{proposition}\vspace{-3mm}

To illustrate the importance of the above result, consider applying the update (\ref{eq:multivariate_copula_updates})  in the case $d = 1$ with $p_{i-1}(x) = \mathbf{1}(x \in [0,1])$; see Section \ref{sec:prac_det} where we initialize at the uniform density. The copula kernel is then $k_{i}(x;x_i) = c_\phi(x,x_i)$, corresponding to Proposition \ref{prop:cop_limits} and Figure \ref{fig:frank_copula}. We see that $k_{i}(1; x_i)$ remains significantly away from 0 for the range of $\phi$ values. On the other hand, even if the observed value $x_i$ is very close to 0 or 1, the Gaussian copula still returns $k_i(0;x_i) = k_i(1;x_i) = 0$, leading to strong boundary bias in the updated $p_i(x)$; see Appendix~\ref{app:additional_experiments} for the practical impact on performance due to this boundary bias.

\subsection{Algorithm} \label{sec:algorithm}

After applying the updates in \eqref{eq:multivariate_copula_updates} to the observed data $x_{1:n}$ to obtain $p_n$, we can generate a posterior sample by predictive resampling $X_i \sim p_{i-1}$ for $i = n+1, \dots, N$ and reapplying the updates, yielding a draw $p_N$. Sampling from MPCP then amounts to combining this draw with an independent sample of the posterior mean event count $\gamma \sim \pi(\gamma\mid n)$ (see Algorithm \ref{alg:mpcp_sampling}). 
We highlight here that Algorithm \ref{alg:mpcp_sampling} does not require MCMC or complex approximate inference techniques like approaches based on GPs.
\begin{algorithm}[t]
\caption{Sampling from MPCP}
\label{alg:mpcp_sampling}
\KwIn{Prior $\pi(\gamma)$,  initial density $p_0$, data $x_{1:n}$}
    \For{$i \leftarrow 1$ \KwTo $n$}{
        Update $p_i \leftarrow \{p_{i-1}, x_i\}$ from \eqref{eq:multivariate_copula_updates} \;
    }
\For{$b \leftarrow 1$ \KwTo $B$}{
    Set $p_n^{(b)} = p_n$\;
    Sample $\gamma^{(b)} \sim \pi(\gamma \mid n)$\;
    \For{$i \leftarrow n+1$ \KwTo $N$}{
        Sample $X_i^{(b)} \sim p_{i-1}^{(b)}$ \;
        Update $p_i^{(b)} \leftarrow \{p^{(b)}_{i-1}, X^{(b)}_i\}$ from \eqref{eq:multivariate_copula_updates} \;
        }
    Compute $\lambda_N^{(b)} = \gamma^{(b)} \times p_N^{(b)}$ \;
    }
\Return $\lambda_N^{(1:B)}$\;
\end{algorithm}

We now formalize convergence of Algorithm \ref{alg:mpcp_sampling}, and in particular show that $\lambda_N = \gamma p_N$ converges under predictive resampling. The Frank copula has an additional key benefit compared to the Gaussian case: we are able to show existence of the limiting intensity $\lambda_\infty$ even in the multivariate setting where $d > 1$, which was not possible for the Gaussian case in \cite{fong2023martingale}. The multivariate setting is of particular importance in the spatial or spatiotemporal setting, where the strengthened result follows from the desirable behavior of the Frank copula at the boundaries. 
\begin{theorem}\label{thm:pred_conv}
Let $\mathcal{X} = [0,1]^d$ for $d \geq 1$, $\phi_j > 0$ for $j = 1,\ldots,d$, and suppose the initial density satisfies $\int_{\mathcal{X}}p_n^2(x)\, \textup{d}x < \infty$. Then under predictive resampling in Algorithm \ref{alg:mpcp_sampling}, there exists a random intensity function $\lambda_\infty \in L^1(\mathcal{X})$ such that
\vspace{-1.5mm}
\begin{align*}
    \int_{\mathcal{X}}\left|\lambda_N(x)-\lambda_\infty(x)\right|\, \textup{d}x \to 0 \quad \textnormal{a.s.}
\end{align*}
\end{theorem}\vspace{-1.5mm}
The strength of the above result relies on the fact that the Frank copula density satisfies $\sup_{u,v}c_\phi(u,v) < \infty$. On the other hand, the Gaussian copula density is poorly behaved at the corners as we have $\lim_{u \to 0} c_\rho(u,u) = \lim_{u \to 1}c_\rho(u,u) = \infty$, which again indicates poor boundary behaviour.

Finally, we highlight the particular interpretability of the posterior mean of MPCP which arises from the factorization (\ref{eq:factorized_posteriors}) and the c.i.d. property. Under the prior {$\pi(\gamma) \propto {\gamma^{-1}}$}, the posterior mean of $\lambda$ satisfies
\begin{align*}
    E[\lambda(x) \mid x_{1:n}] = E[p_\infty(x) \mid x_{1:n}] \, E[\gamma \mid n] = n p_n(x),
\end{align*}
where the first expectation follows from the martingale property, and $p_n$ is the initial density estimated from $x_{1:n}$ given at the start of Algorithm \ref{alg:mpcp_sampling}. This provides a natural connection to frequentist point estimators such as \cite{diggle1985kernel}. For other priors on $\gamma$, we may replace $n$ with the posterior mean under the Poisson-gamma model. More importantly, it guarantees that the centering of the posterior distribution over the mean total count $\gamma$ is not affected by inaccurate estimation of the nonparametric component. This is in contrast to GP-based methods, where the prior on $\lambda$ is induced indirectly through a transformed GP.
 
\subsection{Practical Details}\label{sec:prac_det}

We require an initial density $p_0$ to begin the recursive updates, which allows us to incorporate prior information. By default, we set $p_0$ to the uniform distribution over the input space $p_0 = \mathcal{U}(\mathcal{X})$, corresponding to a homogeneous Poisson process. To estimate the hyperparameters $\phi_1, \dots, \phi_d$, we maximize the prequential log-likelihood $\sum_{i=1}^n \log p_{i-1}(x_i)$ \citep{dawid1984present}. Fitting the MPCP has computational complexity $\mathcal{O}(n^2d)$, since the overhead terms $v_i^j$ for $j = 1, \dots, d$ and $i = 1, \dots, n$ must first be computed via \eqref{eq:multivariate_copula_updates}. Given these terms, evaluating $p_n(x)$ at any point of interest incurs cost $\mathcal{O}(nd)$. 

Since MPCP is sensitive to both the ordering of the observations and the ordering of the covariate dimensions, we follow \citet{fong2023martingale, ghalebikesabi2023quasi} and average predictive distributions over $M=10$ random permutations of the data. We find that the ordering of the covariate dimensions makes little difference (see Appendix~\ref{app:additional_experiments}). A key advantage of MPCP is the ease of predictive resampling. In the univariate case, the update rule \eqref{eq:copula_updates} implies that $p_{i}$ depends on $X_{i}$ only through $P_{i-1}(X_{i})$. Since predictive resampling involves sampling $X_{i} \sim P_{i-1}$, it suffices to draw $P_{i-1}(X_{i}) \sim \mathcal{U}[0,1]$ for $i = n+1, \dots, N$. In the multivariate setting, we similarly only need to simulate $v_{i-1}^{j} \sim \mathcal{U}[0,1]$ for $i = n+1, \dots, N$ and $j = 1, \dots, d$ (see Appendix~\ref{app:notation_preliminaries}). Therefore, drawing a sample $p_N(x)$ has complexity $\mathcal{O}(d(N-n))$.

\section{Experiments} \label{sec:experiments}

The code is available at \url{https://github.com/joshuarooijakkers/mpcp}. We evaluate our approach on both synthetic and real-world data, comparing it with several existing methods. Specifically, we consider GP-based Cox processes with the exponential link function \citep{moller1998log}, fitted using both MCMC (LGCP) and variational inference (VLGCP), as well as the squared link function fitted using variational inference (VBPP; \citealp{lloyd2015variational}). We further compare against the DPMM of \citet{kottas2006dirichlet}. All methods are run on a GPU, except for the DPMM, which does not benefit from parallelization and is therefore run on a CPU. For MPCP, we use the prior {$\pi(\gamma) \propto {\gamma^{-1}}$} as described in \Cref{sec:predictive_resampling_cox} and set $N - n= 2000$ for the number of future samples, which we find to be sufficiently large for convergence (see Appendix~\ref{app:additional_experiments}). Further details on the experimental setup are provided in Appendix~\ref{app:additional_experiments}.

\subsection{Synthetic Data} \label{sec:synthetic_data}

We first evaluate our method on two simulated datasets in one and two dimensions. Specifically, we consider the following intensity functions:
\begin{align*}
    \lambda_1(x) &= 2 \exp (-x/15) + \exp(-((x-25)/10)^2), \\
    \lambda_2(x,x') &= 2\cos^5\!(\pi(x-x')/20) \!+\! 0.2\sin^2\!(\pi(x+x')/10)
\end{align*}
with domains $\mathcal{X}_1 = [0,50]$ and $\mathcal{X}_2 = [0,10]^2$ for $\lambda_1$ and $\lambda_2$, respectively. Following the literature, we quantitatively evaluate the methods using the squared error $L_2 := |\mathcal{X}|^{-1} \int_\mathcal{X} (\lambda^*(x) - \overline{\lambda}(x))^2 \textup{d}x$ and the joint test log-likelihood $L_\textup{test} := \log p(x_{1:n_\textup{test}}^{\textup{test}} \mid x_{1:n})$
 where $\lambda^*$ denotes the true intensity, $\overline{\lambda}(x) := \mathbb{E}[\lambda(x)\mid x_{1:n}]$ is the posterior mean intensity, and $x_{1:n_{\mathrm{test}}}^{\mathrm{test}} = \{x^{\mathrm{test}}_i\}_{i=1}^{n_\textup{test}}$ denotes the test points. The results are reported in \Cref{tab:synthetic_data} and we plot the posterior estimates in \Cref{fig:1D_comparison} for $\lambda_1$ and in Appendix~\ref{app:additional_experiments} for $\lambda_2$. We note that the DPMM approach does not have a publicly available two-dimensional implementation and is therefore only evaluated in one dimension. 

\begin{figure*}[t]
  \centering
  \includegraphics[width=\textwidth]{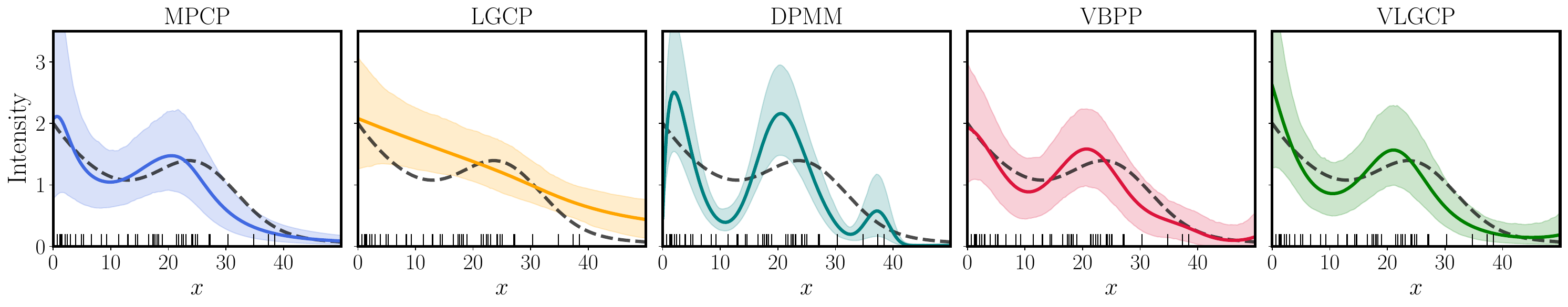}
  \caption{Posterior mean intensity and corresponding 90\% credible interval, together with the observed events (vertical ticks) generated from the true intensity function $\lambda_1$ (dashed).}
  \label{fig:1D_comparison}
\end{figure*}

We see that the $L_2$ performance (which evaluates only the mean intensity) of MPCP is reasonably competitive with that of the GP-based approaches, although these methods outperform MPCP. However, for the test log-likelihood, which also accounts for uncertainty, MPCP outperforms all competing approaches while requiring only a fraction of their computational cost. In particular, inference for MPCP is orders of magnitude faster than the MCMC-based approaches and is also faster than the variational approximations used for the GP-based methods. As we will see in the real-data analysis, this computational gap increases further as the sample size $n$ increases, due to the poor scaling of GP-based approaches. The DPMM performs noticeably worse than the other approaches, likely due to hyperparameter choice being challenging without a tractable marginal likelihood.

\begin{table}[h]
    \centering
    \captionof{table}{Mean and standard error of squared error $L_2$,
        test log-likelihood $L_\textup{test}$ and runtime across 50 seeds
        for synthetic intensity functions $\lambda_1$ and $\lambda_2$. Methods are grouped into MCMC-based methods, approximate inference methods, and our proposed method, from top to bottom.}
        \label{tab:synthetic_data}

        \resizebox{0.9\textwidth}{!}{
        \begin{tabular}{lcccccc}
        \toprule
        & \multicolumn{3}{c}{$\lambda_1$ $(d=1, n \approx 46)$}
        & \multicolumn{3}{c}{$\lambda_2$ $(d=2, n \approx 118)$} \\
        & $L_2$ & $L_\textup{test}$ & Runtime\,(s)
        & $L_2$ & $L_\textup{test}$ & Runtime\,(s) \\
        \cmidrule(lr){2-4} \cmidrule(lr){5-7}
        LGCP  & $\bm{0.10 \pm 0.01}$ & $-34.3 \pm 1.06$ & $92.3 \pm 0.84$
               & $\bm{0.11 \pm 0.01}$ & $-54.1 \pm 1.82$ & $138 \pm 0.26$ \\
        DPMM  & $0.28 \pm 0.03$ & $-46.7 \pm 0.51$ & $62.2 \pm 1.91$
               & -- & -- & -- \\
        \midrule
        VLGCP & $0.12 \pm 0.01$ & $-22.5 \pm 0.33$ & $0.95 \pm 0.03$
               & $0.14 \pm 0.01$ & $-38.2 \pm 0.59$ & $14.4 \pm 0.59$ \\
        VBPP  & $0.11 \pm 0.01$ & $-23.9 \pm 0.36$ & $2.36 \pm 0.37$
               & $0.12 \pm 0.01$ & $-39.4 \pm 0.54$ & $19.0 \pm 0.64$ \\
        \midrule
        MPCP  & $0.13 \pm 0.01$ & $\bm{-22.1 \pm 0.25}$ & $\bm{0.49 \pm 0.07}$
               & $0.17 \pm 0.02$ & $\bm{-38.0 \pm 1.15}$ & $\bm{4.01 \pm 0.05}$ \\
        \bottomrule
        \end{tabular}
        }
\end{table}

\begin{figure}[t]
    \centering
        \includegraphics[width=0.8\textwidth]{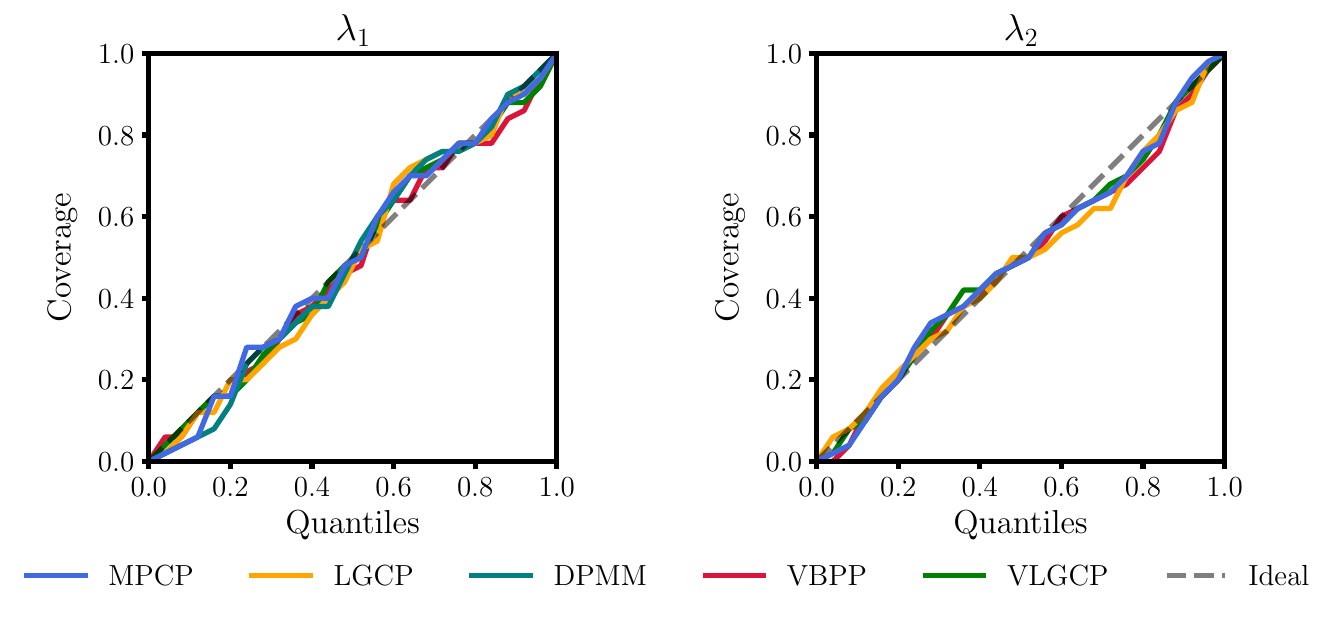}
        \captionof{figure}{{Coverage plots of the $q$-level credible intervals for  $\gamma = \int_{\mathcal{X}} \lambda(x) \textup{d}x$}.}
        \label{fig:coverage_main}
\end{figure}

We see in \Cref{fig:1D_comparison} that the MPCP provides comparable uncertainty quantification on $\lambda$ compared to the other approaches. We further assess the uncertainty through coverage plots of $\gamma = \int_{\mathcal{X}} \lambda(x)\, \textup{d}x$ in \Cref{fig:coverage_main}, where we see that all approaches achieve good coverage performance. This is unsurprising for the MPCP, as the posterior over $\gamma$ is the well-understood conjugate Gamma-Poisson model. In Appendix~\ref{app:additional_experiments}, we also find that the interval lengths are similar between all approaches. These results highlight the reliability of the uncertainty quantification provided by MPCP.

\subsection{Real Data}\label{sec:realdata}

Finally, we evaluate MPCP on two larger-scale real-world datasets. We first consider the \textsc{bei} dataset, which records the locations of trees in the tropical rainforest on Barro Colorado Island \citep{hubbell1983diversity}. The locations are two-dimensional with 3,604 observed events in total. We additionally consider a spatio-temporal \textsc{earthquake} dataset obtained from the U.S. Geological Survey Earthquake Catalog~\citep{usgs_earthquake_catalog}. This dataset consists of earthquakes with Richter magnitude greater than 4 in and around Japan over the 10-year period from 2000 to 2009. Incorporating both spatial location and time yields a three-dimensional intensity, with 8,954 observed events in total.

\begin{figure*}[h!]
  \centering
  \includegraphics[width=\textwidth]{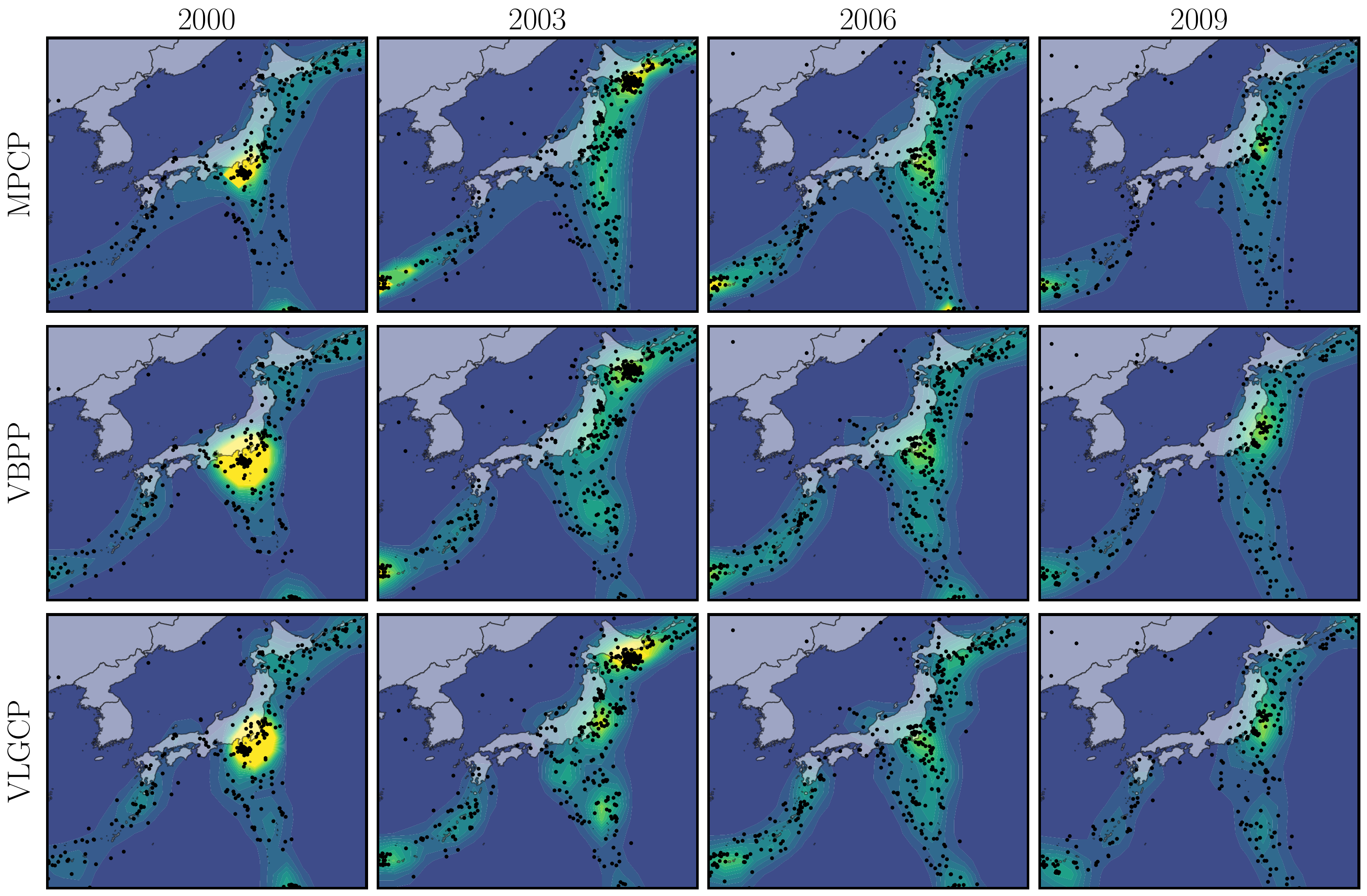}
  \caption{Posterior mean intensity over time for {\textsc{earthquake}} data for MPCP, VBPP, and VLGCP (top to bottom) at four time points (left to right), with black points indicating the observed earthquake events. Lighter colors indicate higher mean intensity.}
  \label{fig:eq_mean}
\end{figure*}

In the absence of a true intensity, we follow the literature \citep{aglietti2019structured, sellier2023sparse} and split each dataset into training and test sets of equal size. Thus, the number of training observations is $n=1802$ and $n=4477$ for \textsc{bei} and \textsc{earthquake} datasets, respectively. We then evaluate the predictive performance using the test log-likelihood. At these scales, the MCMC-based methods are prohibitively expensive to run, and we therefore restrict the comparison to variational inference methods. The results are reported in \Cref{tab:real_data}, and a comparison of the posterior means for the \textsc{earthquake} dataset is shown in \Cref{fig:eq_mean}. Posterior mean plots for the \textsc{bei} dataset, together with posterior variance plots for both datasets, are provided in Appendix~\ref{app:additional_experiments}.

\begin{figure}[h]
  \centering
  \includegraphics[width=0.6\textwidth]{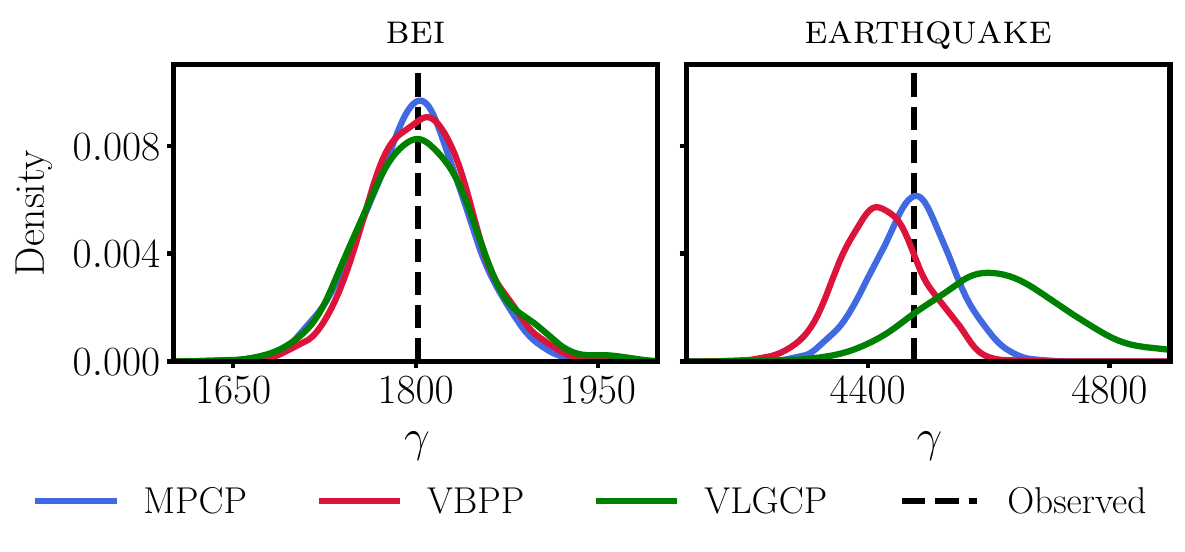}
  \caption{Posterior density of $\gamma = \int \lambda(x)\,\textup{d}x$, together with the true observed count $n$ (dashed).}
  \label{fig:counts}
\end{figure}

\begin{table}[h]
\centering
\caption{Mean and standard error of the test log-likelihood $L_\textup{test}$ and runtime across 10 seeds for \textsc{bei} and \textsc{earthquake} datasets.} \label{tab:real_data}
\resizebox{0.7\columnwidth}{!}{%
\begin{tabular}{lcccc}
\toprule
& \multicolumn{2}{c}{\textsc{bei}}
& \multicolumn{2}{c}{\textsc{earthquake}} \\
& $L_\textup{test}$ & Runtime (s)
& $L_\textup{test}$ & Runtime (s) \\
\cmidrule(lr){2-3}\cmidrule(lr){4-5}
VLGCP & $-10958 \pm 8.68$ & $44.5 \pm 2.40$ & $-51.4 \pm 26.4$ & $300 \pm 8.55$ \\
VBPP  & $-10929 \pm 12.0$ & $77.0 \pm 4.63$ & $-196 \pm 13.7$ & $911 \pm 45.2$ \\
\midrule
MPCP  & $\bm{-10778 \pm 10.3}$ & $\bm{21.9 \pm 0.97}$ & $\bm{1094 \pm 64.3}$ & $\bm{48.6 \pm 8.76}$ \\
\bottomrule
\end{tabular}
}
\end{table}

 We see in \Cref{tab:real_data} that MPCP achieves the best test log-likelihood and requires the shortest runtime for both datasets, where the differences in predictive performance and runtime are particularly large for the \textsc{earthquake} dataset. The noticeably worse performance of the GP-based methods in \textsc{earthquake} is primarily due to the three-dimensional input space, which poses computational challenges for the sparse approximations required by the variational methods. Due to memory constraints, we were limited to 10 inducing points/bins per dimension (i.e. $10^3$ total points), as larger configurations exceeded available memory, and drastically increased the already large run-times. In Appendix \ref{app:additional_experiments}, we verify through ablation studies that this performance gap is indeed driven by the sparse approximation, highlighting a strong advantage of the (almost) exact inference approach of MPCP.   In \Cref{fig:eq_mean}, we see that MPCP successfully captures the changing intensity over time, with posterior mean estimates broadly similar to those of the other GP-based approaches. Around the year 2000, however, the GP-based approaches oversmooth the high intensity region relative to the data, likely due to the limited number of inducing points/bins.

To evaluate the uncertainty quantification in both datasets, we also plot the posterior distribution of $\gamma = \int_{\mathcal{X}} \lambda(x)\, \textup{d}x$ in \Cref{fig:counts}, with the observed sample size $n$ in dashed. As alluded to in Section \ref{sec:algorithm}, the posterior of MPCP for $\gamma$ is centered at $n$ due to the factorization. In contrast, the GP-based posteriors for $\gamma$ exhibit bias away from the observed $n$ for the \textsc{earthquake} dataset, which likely arises due to poor variational approximations as also indicated in Table \ref{tab:real_data} and Appendix \ref{app:additional_experiments}.  

\section{Conclusion}

In this paper, we introduce the Martingale Posterior Cox Process (MPCP) based on a copula-based updating rule that is particularly well suited to multivariate and bounded domains. Unlike other nonparametric approaches in the literature, MPCP does not require a complex or difficult-to-interpret prior construction, while still allowing for (almost) exact inference and accurate uncertainty quantification without MCMC or variational inference. As a result, MPCP avoids the long runtimes associated with MCMC, as well as the approximation error and poor scaling with input dimensionality that can arise with variational inference for GP-based methods. MPCP inherits the flexibility of KDE approaches, as demonstrated by strong predictive performance, while returning Bayesian uncertainty quantification. 
An interesting future direction is to extend the MPCP framework to other classes of point processes, such as Cox-Hawkes processes \citep{miscouridou2023coxhawkes} and marked point processes \citep{taddy2012mixture}. Another avenue is to further accelerate MPCP and reduce its $O(n^2d)$ computational complexity through suitable approximations.

\subsubsection*{Acknowledgements}
Edwin Fong receives funding from the Research Grants Council of
Hong Kong through the Early Career Scheme (Grant 27304424) and the General Research Fund (Grant 17306925). 

\bibliographystyle{apalike}
\bibliography{references}

\appendix
\newpage
\appendixtitle


\section{Preliminaries} \label{app:notation_preliminaries}
\subsection{Martingale Posterior Distributions}
We briefly outline some details regarding the construction of the general martingale posterior. Under the c.i.d. condition, since the CDF $P_i(x)$ is a martingale, \cite{berti2004limit} established the existence of a random distribution $P_\infty$ such that 
$$P_N \to P_\infty$$
weakly a.s. as $N \to \infty$. Furthermore, the above is usually enough to imply
\begin{align*}
    \frac{1}{N}\left(\sum_{i = 1}^n \delta_{x_i} + \sum_{j = n+1}^N \delta_{X_j}\right) \to P_\infty
\end{align*}
weakly a.s. as $N \to \infty$. In more general settings, the a.s. weak convergence can be established directly through other means, usually through the identification of other (approximate) martingales, e.g. \cite{battiston2025bayesian,fong2025bayesian,cui2026bayesian}. 

The martingale posterior is defined as the distribution of a functional $\theta(\cdot)$ of the limiting random measure $P_\infty$, that is
$$\theta(P_\infty)\sim \pi(\theta \mid x_{1:n}).$$
Since the limiting empirical measure has the same weak limit, we may also write $\theta(x_{1:n}, X_{n+1:\infty})$. In practice, we will require a finite but sufficiently large truncation $N$, so we will work with the distribution of $\theta(P_N)$ or $\theta(x_{1:n}, X_{n+1:N})$. In some cases, the object of interest may also be nonparametric, such as a mixing distribution in \cite{Fortini2020}, the quantile function in \cite{fong2025bayesian}, or the intensity function in our setting. This latter case is particularly delicate, as it is not guaranteed that $P_\infty$ has a probability density function, e.g. \citet{berti2013exchangeable}. Theorem \ref{thm:pred_conv} establishes the existence of $p_\infty$ and thus $\lambda_\infty$ in our setting. 

\subsection{Bivariate Copula Updates}
In this section, we outline additional details regarding the bivariate copula update within the martingale posterior framework. Let $x = (x^1,\ldots, x^d)$ in the multivariate case. Following \cite{fong2023martingale}, we may integrate (\ref{eq:multivariate_copula_updates}) with respect to $\{x^d, \ldots, x^{k+1}\}$ in descending order, which gives the following recursive update for the conditional CDFs $u_i^k = P_i(x^k \mid x^{1:k-1})$ for a grid point $x \in \mathbb{R}^d$:
\begin{align}\label{eq:copula_cdf_mv}
    u_{i}^k = \frac{(1-\alpha_{i})u_{i-1}^k + \alpha_i H_{\phi_k}(u_{i-1}^k, v_{i-1}^k) \prod_{j = 1}^{k-1}c_{\phi_j}(u_{i-1}^j, v_{i-1}^j)}{1-\alpha_{i} + \alpha_i \prod_{j = 1}^{k-1}c_{\phi_j}(u_{i-1}^j, v_{i-1}^j)}
\end{align}
for $k = 1,\ldots, d$. Note that $v_{i-1}^k = P_{i-1}(x_i^k \mid x_i^{1:k-1})$ can be computed from the above formula as well, where $x_i$ is the observed datum. 

Under predictive resampling, the copula update has a particularly nice property. Since $X_{i+1} \sim P_i$, the standard integral probability transform gives $V_{i}^k = P_i(X_{i+1}^k \mid X_{i+1}^{1:k-1}) \sim \mathcal{U}[0,1]$ independently across $k = 1,\ldots,d$. As a result, predictive resampling involves applying (\ref{eq:multivariate_copula_updates}) and (\ref{eq:copula_cdf_mv}) to uniformly generated random variables. 

\section{Proofs of Theoretical Results} \label{app:proofs_theoretical_results}

\subsection{Proof of Proposition \ref{prop:cop_limits}}
The bivariate Gaussian copula density has the form
\begin{align*}
    c_\rho(u,v) = \frac{1}{\sqrt{1-\rho^2}}\exp\left\{- \frac{\rho^2(z_u^2 + z_v^2) - 2\rho z_u z_v  }{2(1-\rho^2)}\right\}
\end{align*}
where $z_u = \Phi^{-1}(u), z_v = \Phi^{-1}(v)$. We can complete the square for the exponent, giving
\begin{align*}
    z_u^2 + z_v^2 - 2\frac{z_u z_v}{\rho} = \left(z_u - \frac{z_v}{\rho}\right)^2 - \frac{1-\rho^2}{\rho^2} z_v^2,
\end{align*}
which when substituted into the Gaussian copula density gives
\begin{align*}
    c_\rho(u,v) = \frac{\exp\left({z_v^2}/2\right)}{\sqrt{1-\rho^2}} \exp\left\{- \frac{\rho^2}{2(1-\rho^2)}\left(z_u - \frac{z_v}{\rho}\right)^2\right\}.
\end{align*}
For fixed $v \in (0,1)$ and $\rho \in (0,1)$, taking $u \to 0$ or $u \to 1$ gives $z_u^2 \to \infty$, so we have $\lim_{u \to 0}c_\rho(u,v)= \lim_{u \to 1}c_\rho(u,v) = 0$. 

For the result on the Frank copula, we note that
\begin{align*}
    \lim_{u \to 0} c_\phi(u, v) &= \frac{-\phi (e^{-\phi} - 1) e^{-\phi(0 + v)}}{(e^{-\phi} - 1 + (e^{-\phi 0} - 1)(e^{-\phi v} - 1))^2} = \frac{\phi e^{-\phi v}}{1-e^{-\phi} }\cdot
\end{align*}
The same argument can be applied for $u \to 1$. 

\subsection{Proof of Theorem \ref{thm:pred_conv}}

To begin, we have the strong upper bound below.
\begin{lemma}
For $\phi > 0$, the bivariate Frank copula density satisfies
   \begin{align*}
    \sup_{u,v} c_\phi(u,v) \leq \frac{\phi}{1-e^{-\phi}} < \infty.
\end{align*} 
\end{lemma}
\begin{proof}
To see this, let $x = e^{-\phi u}$ and $y = e^{-\phi v}$, so we can write
\begin{align*}
    c_\phi(u,v) = \frac{\phi(1-e^{-\phi}) \, xy}{\{1-e^{-\phi} - (1-x)(1-y)\}^2}
\end{align*}
where $e^{-\phi} \leq x,y \leq 1$. First, we note that both the numerator and the term in the bracket in the denominator are positive. Next, for the denominator, we note that the AM-GM inequality gives $x + y \geq 2\sqrt{xy}$, which gives
\begin{align*}
(1-x)(1-y) =  1- (x+y) + xy \leq (1-\sqrt{xy})^2.
\end{align*}
Next, we note that $\sqrt{xy} \geq e^{-\phi}$, so 
\begin{align*}
  (1-\sqrt{xy})^2 \leq (1-e^{-\phi})(1-\sqrt{xy}).
\end{align*}
Together, this gives
\begin{align*}
    1-e^{-\phi} - (1-x)(1-y) \geq (1-e^{-\phi})\left\{1 - (1-\sqrt{xy})\right\} = (1-e^{-\phi})\sqrt{xy}.
\end{align*}
Together, this gives
\begin{align*}
    c_\phi(u,v) \leq \frac{\phi(1-e^{-\phi}) xy}{(1-e^{-\phi})^2 xy} = \frac{\phi}{1 - e^{-\phi}}\cdot
\end{align*}
\end{proof}
From the above lemma, we have
\begin{align*}
q_\phi(u):= \int_0^1 c_\phi(u,v)^2 \, dv  &\leq \frac{\phi}{1-e^{-\phi}} \int_0^1 c_\phi(u,v)\, dv \\
 &= \frac{\phi}{1-e^{-\phi}} < \infty.
\end{align*}
Unlike the Gaussian case,  $q_\phi(u)$ is uniformly bounded over all $u \in [0,1]$.

We now follow a similar argument to \citet[Theorem 5]{fong2023martingale}, where we will apply \citet[Theorem 4]{berti2013exchangeable} to show the limiting measure $P_\infty$, which exists from the c.i.d. condition, has a Lebesgue density a.s. A sufficient condition is that
\begin{align}\label{eq:bounded_pn2_condit}
    \sup_{N \geq n+1} E\left[\int_K p_N(x)^2\, \textup{d}x \mid x_{1:n}\right] < \infty \quad \text{a.s.}
\end{align}
for all compact subsets $K \subseteq \mathcal{X}\subseteq \mathbb{R}^d$. To compute the above, we can first calculate the below expectation using the recursion (\ref{eq:multivariate_copula_updates}):
\begin{align*}
    E[p_{N}(x)^2 \mid x_{1:n},X_{n+1:N-1}]  &= p_{N-1}(x)^2\left\{(1-\alpha_{N})^2 + 2\alpha_N(1-\alpha_N)E_{V^{1:d}}\left[\prod_{j  =1}^d c_{\phi_j}(u_{N-1}^j, V^j) \right] \right. \\
    &\left. \quad  +\alpha_N^2 \,E_{V^{1:d}}\left[\prod_{j  =1}^d c_{\phi_j}(u_{N-1}^j, V^j)^2\right]\right\}.
\end{align*}
Since $V^j \sim \mathcal{U}[0,1]$ independently across $j = 1,\ldots, d$ and $E_V[c_\phi(u,V)] = 1$, the above simplifies to
\begin{align*}
     E[p_{N}(x)^2 \mid x_{1:n},X_{n+1:N-1}]  &= p_{N-1}(x)^2\left\{1- \alpha_N^2 + \alpha_N^2 \prod_{j = 1}^d q_{\phi_j}(u_{N-1}^j) \right\}\\
     &\leq p_{N-1}(x)^2\left(1+ C\alpha_N^2\right)
\end{align*}
where
\begin{align*}
    C = \prod_{j = 1}^d \frac{\phi_j}{1-e^{-\phi_j}} - 1 < \infty.
\end{align*}
Since $\alpha_N = O(N^{-1})$, we have that $\prod_{i = n+1}^\infty (1+ C\alpha_i^2) < \infty$, so there exists another finite constant $D$ independent of $N$ and $x$ such that
\begin{align*}
    \sup_{N \geq n+1} E[p_N(x)^2 \mid x_{1:n}] \leq D p_n(x)^2. 
\end{align*}
Integrating the above over $x$, applying Tonelli's theorem, and using the assumption $\int_{\mathcal{X}} p_n^2(x)\, \textup{d}x < \infty$, we thus have that (\ref{eq:bounded_pn2_condit}) holds.

Since each predictive distribution has a density $p_N$, \citet[Theorem 4]{berti2013exchangeable} thus gives that $P_\infty$ has a Lebesgue density $p_\infty$. Furthermore, \citet[Theorem 1]{berti2013exchangeable} then gives
\begin{align*}
 \int_{\mathcal{X}} |p_N(x) - p_\infty(x)| \, \textup{d}x \to 0 \quad \text{a.s.}
\end{align*}
Finally, letting $\lambda_N = \gamma p_N$ and $\lambda_\infty = \gamma p_\infty$ where $\gamma \sim \pi(\gamma \mid n)$ and $\gamma < \infty$ a.s., we have 
\begin{align*}
    \int_{\mathcal{X}} |\lambda_N(x) - \lambda_\infty(x)| \, \textup{d}x = \gamma\int_{\mathcal{X}}\left| p_N(x) - p_\infty(x)\right|\, \textup{d}x, 
    \end{align*}
    so we have the final result. 

\section{Additional Experiments} \label{app:additional_experiments}

Experiments were run on a 2025 HP ZBook Power G11 Mobile Workstation PC with an Intel(R) Core(TM) Ultra 7 155H CPU with 16GB memory. MPCP, LGCP, VLGCP and VBPP were run on an NVIDIA RTX 3000 GPU with 16GB memory.

\subsection{Alternative Copula Choices}
\begin{figure}[h!]
    \centering
    \begin{subfigure}[b]{0.49\textwidth}
        \centering
        \includegraphics[width=\textwidth]{images/methodology/copula_update_narrow.pdf}
        \caption{Frank copula for $\phi = 1, 5, 10$ (\textcolor[RGB]{198, 219, 239}{\rule[0.5ex]{0.3cm}{1.5pt}},\textcolor[RGB]{107, 174, 214}{\rule[0.5ex]{0.3cm}{1.5pt}},\textcolor[RGB]{33, 113, 181}{\rule[0.5ex]{0.3cm}{1.5pt}}).}
        \label{fig:frank_copula_update_app}
    \end{subfigure}
    \hspace{0.001\textwidth}
    \begin{subfigure}[b]{0.49\textwidth}
        \centering
        \includegraphics[width=\textwidth]{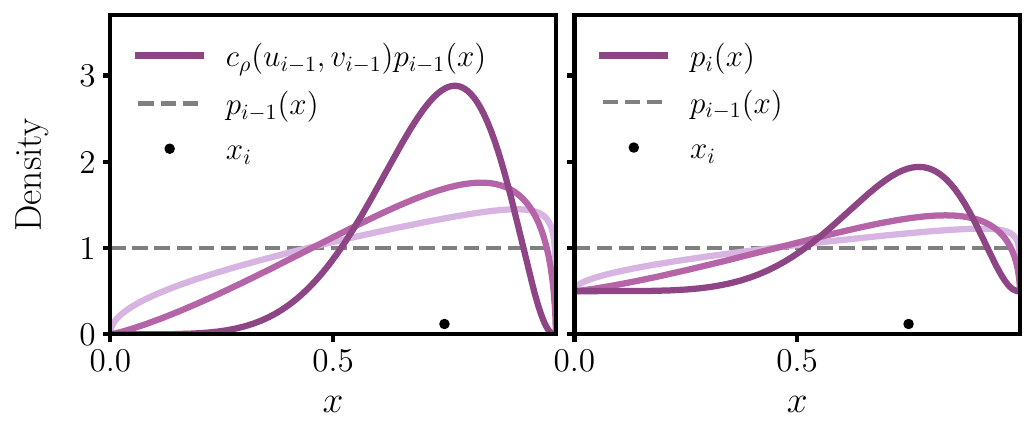}
        \caption{Gaussian copula for $\rho = 0.5, 0.7, 0.9$ 
        (\textcolor[RGB]{216, 180, 226}{\rule[0.5ex]{0.3cm}{1.5pt}},
        \textcolor[RGB]{181, 101, 167}{\rule[0.5ex]{0.3cm}{1.5pt}},
        \textcolor[RGB]{142, 69, 133}{\rule[0.5ex]{0.3cm}{1.5pt}}).}
        \label{fig:gaussian_copula_update_app}
    \end{subfigure}
    \caption{Comparison of predictive updates under the Frank and Gaussian copulas. For each copula, the left panel shows the copula kernel
$k_i(x;x_i) := c(P_{i-1}(x), P_{i-1}(x_i))p_{i-1}(x)$, while the right panel shows the resulting predictive density $p_i(x)$, with $x_i = 0.75$ and $\alpha_i = 0.5$.}
    \label{fig:kernel_comparison_MPCP}
\end{figure}

We now empirically motivate the choice of the Frank copula, as discussed in \Cref{sec:methodology}, by comparing it with the Gaussian copula used in \citet{hahn2018recursive, fong2023martingale}. First, we visualize the copula update as an online kernel density estimator with kernel $k_i(x;x_i) := c(P_{i-1}(x), P_{i-1}(x_i))p_{i-1}(x)$ for both the Frank copula in \Cref{fig:frank_copula_update_app} and Gaussian copula in \Cref{fig:gaussian_copula_update_app}. We observe that, unlike the Frank copula, the Gaussian copula causes the proposal kernel $k_i(x;x_i)$ to vanish at the boundaries. This follows directly from the boundary limits derived in Proposition~\ref{prop:cop_limits}.

\begin{figure}[h!]
  \centering
  \includegraphics[width=0.6\textwidth]{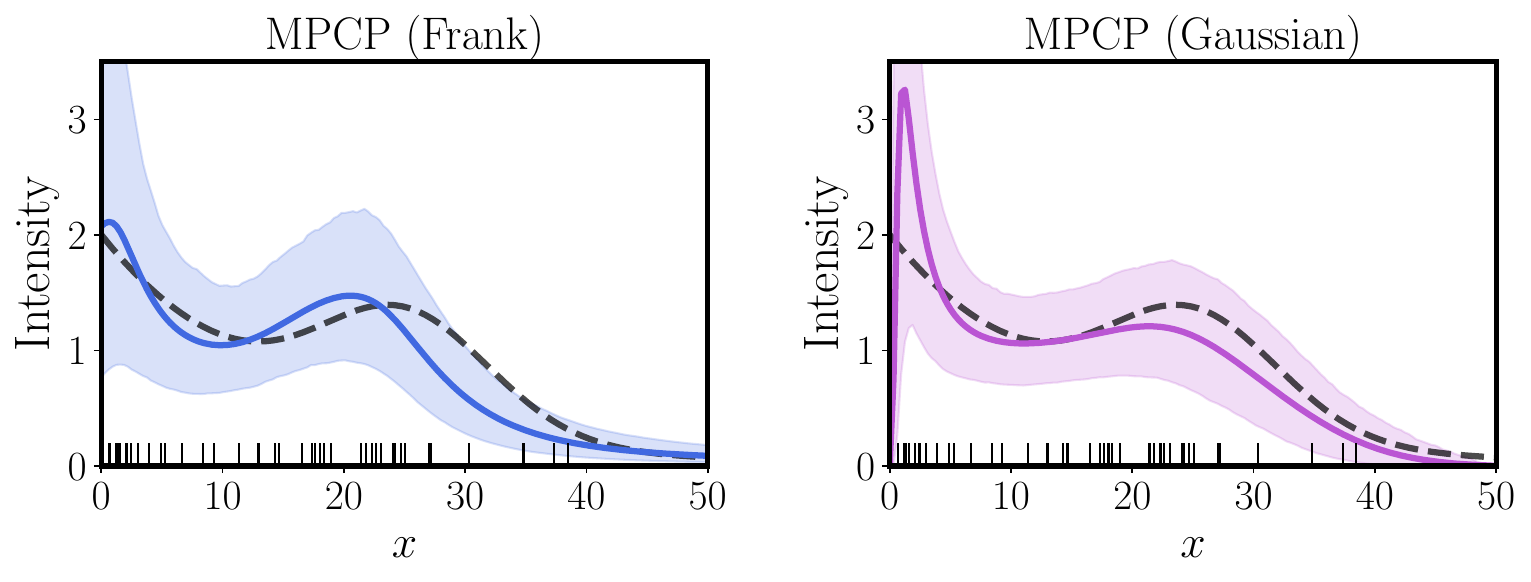}
  \caption{Posterior mean intensity and corresponding 90\% credible interval for MPCP with the Frank (left) and Gaussian (right) copula. The observed events (vertical ticks) are generated from the true intensity function $\lambda_1$ (dashed).}
  \label{fig:MPCP_copula_comparison}
\end{figure}

Next, we examine how this boundary bias affects the performance of MPCP on the 1D synthetic dataset described in \Cref{sec:synthetic_data}, with the results shown in \Cref{fig:MPCP_copula_comparison}. For the Gaussian copula, the estimated intensity is strongly biased downwards at the boundaries because the corresponding proposal kernel has zero density there, as illustrated in \Cref{fig:kernel_comparison_MPCP}. At the left boundary, this leads to a less accurate estimate: despite the high concentration of observed events near the boundary, the estimated intensity has a sharp increase immediately before reaching near-zero at the boundary itself. In contrast, the Frank copula does not exhibit this boundary bias and provides a more accurate estimate near the boundaries.

\subsection{Convergence of Predictive Resampling}
\begin{figure}[h!]
  \centering
  \includegraphics[width=0.9\textwidth]{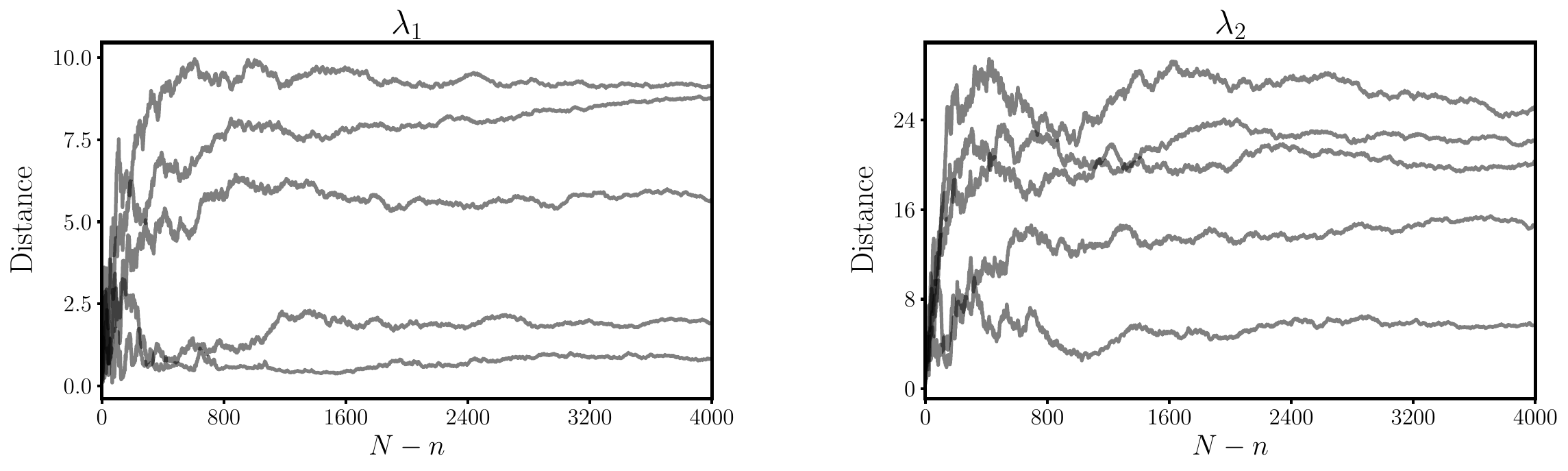}
  \caption{Trajectories of $L_N$, as defined in \eqref{eq:L}, over $N-n$ predictive resampling steps for different posterior samples in the synthetic data examples.}
  \label{fig:convergence}
\end{figure}

To illustrate the convergence of predictive resampling, as established by Theorem \ref{thm:pred_conv}, we consider trajectories of
\begin{align}\label{eq:L}
L_N = \int_{\mathcal{X}} \left|\lambda_N(x) - \lambda_n(x)\right|\textup{d}x,
\end{align}
which measures the $L^1$ distance between the intensity after $N-n$ predictive resampling steps, $\lambda_N$, and the initial intensity, $\lambda_n$. Since the limiting intensity $\lambda_\infty$ is not available in practice, this quantity provides a convenient proxy for monitoring convergence. We find in Figure~\ref{fig:convergence} that $N-n=2000$ predictive resampling steps is sufficient for convergence.

\subsection{Predictive Uncertainty for Density Component}

\begin{figure}[h]
  \centering
  \includegraphics[width=\textwidth]{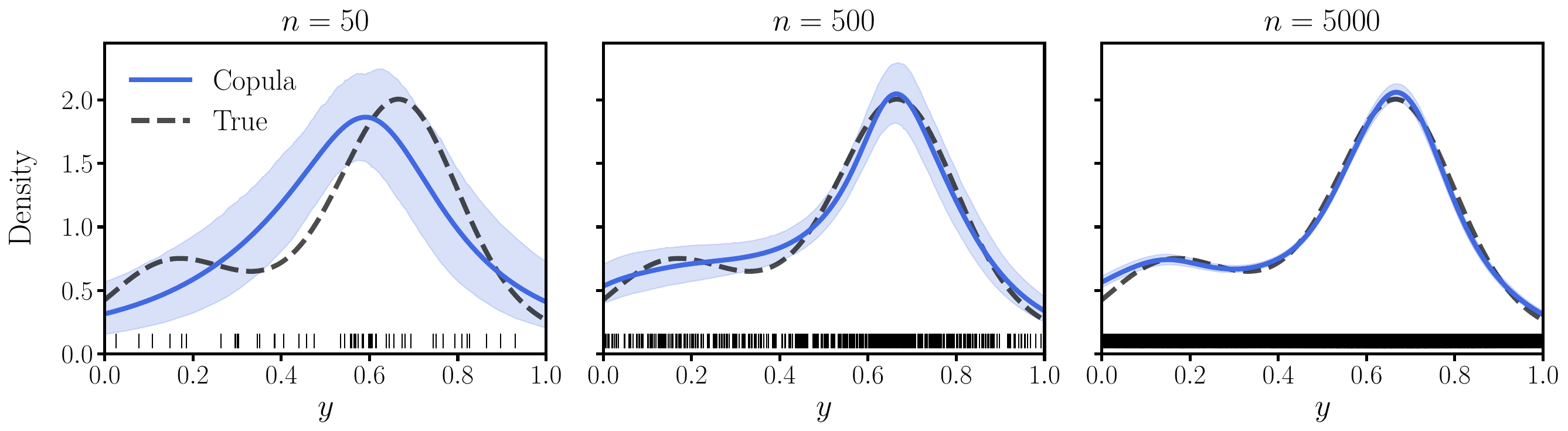}
  \caption{Density component of MPCP for $n=50$ (left), $n=500$ (middle), and $n=5,000$ (right), together with the true density (dashed).}
  \label{fig:pred_uncert}
\end{figure}

In \Cref{sec:methodology}, we show that the posterior intensity decomposes into a density and a count component, with the uncertainty in the density component decreasing as more events are observed. We now illustrate this behavior through predictive resampling. Specifically, we simulate from a 20-component Gaussian mixture and normalize it to the unit interval, resulting in a multimodal intensity on $\mathcal{X} = [0,1]$. We then generate an increasing number of events from this intensity and fit the density component of MPCP, and the resulting posterior densities are shown in \Cref{fig:pred_uncert}. As the number of observed events increases, the posterior density concentrates around the true density and the predictive uncertainty decreases. 

\subsection{Covariate Ordering of MPCP}

Since MPCP is sensitive to the ordering of the covariate dimensions, we investigate the effect of this ordering on its performance. Specifically, we consider the 2D synthetic experiment described in \Cref{sec:synthetic_data} and vary the order in which the covariate dimensions are processed. The results are reported in \Cref{tab:change_order}, including the 90\% coverage and corresponding interval length for $\gamma = \int_\mathcal{X}\lambda(x)\textup{d}x$. We find that the standard errors across the 50 random seeds overlap substantially across the different orderings, suggesting that the ordering of the covariate dimensions has little effect on the overall performance of MPCP.

\begin{table}[h!] 
\centering 
\caption{Mean and standard error across 50 seeds for $L_2$, $L_\textup{test}$, 90\% coverage, and the corresponding 90\% interval length, under different orderings of the covariate dimensions.} \label{tab:change_order} 
\resizebox{0.5\textwidth}{!}{
\begin{tabular}{lcc} 
\toprule 
& $(x,x')$ & $(x', x)$ \\ \midrule 
$L_2$ & $0.17 \pm 0.02$ & $0.16 \pm 0.02$ \\ $L_\textup{test}$ & $-38.0 \pm 1.15$ & $-37.4 \pm 1.06$ \\ 
90\% coverage & 0.90 & 0.90 \\ 
90\% interval length & $35.4 \pm 0.27$ & $35.4 \pm 0.27$ \\ \bottomrule 
\end{tabular}
}
\end{table}

\subsection{Synthetic Experiments} \label{app:synthetic_experiments}

We evaluate MPCP and competing methods on data simulated from the following one and two-dimensional intensity functions, respectively:
\begin{align*}
    \lambda_1(x) &= 2 \exp (-x/15) + \exp(-((x-25)/10)^2), \qquad \qquad \quad\mathcal{X}_1 = [0, 50], \\
    \lambda_2(x,x') &= 2 \cos^5(\pi (x - x')/20) + 0.2 \sin^2(\pi (x + x')/10), \qquad \mathcal{X}_2 = [0,10]^2.
\end{align*}
For MPCP, we use the setup described in \Cref{sec:methodology} with the prior {$\pi(\gamma) \propto {\gamma^{-1}}$}. For all GP-based approaches, we use a squared exponential kernel and a constant mean function. LGCP and VLGCP approximate the intensity through spatial discretization into bins \citep{moller1998log}, using 20 bins for $\lambda_1$ and $12^2 = 144$ bins for $\lambda_2$. VBPP \citep{lloyd2015variational} instead uses inducing points; we use 20 inducing points for $\lambda_1$ and $12^2 = 144$ inducing points for $\lambda_2$. We found that increasing the number of bins or inducing points did not lead to a significant improvement in performance on the synthetic experiments. VLGCP and VBPP use variational approximations, whereas LGCP uses Hamiltonian Monte Carlo. Finally, the DPMM uses a mixture of beta kernels and the MCMC method of \cite{neal2000markov}. 

\begin{figure}[h]
  \centering
  \includegraphics[width=\textwidth]{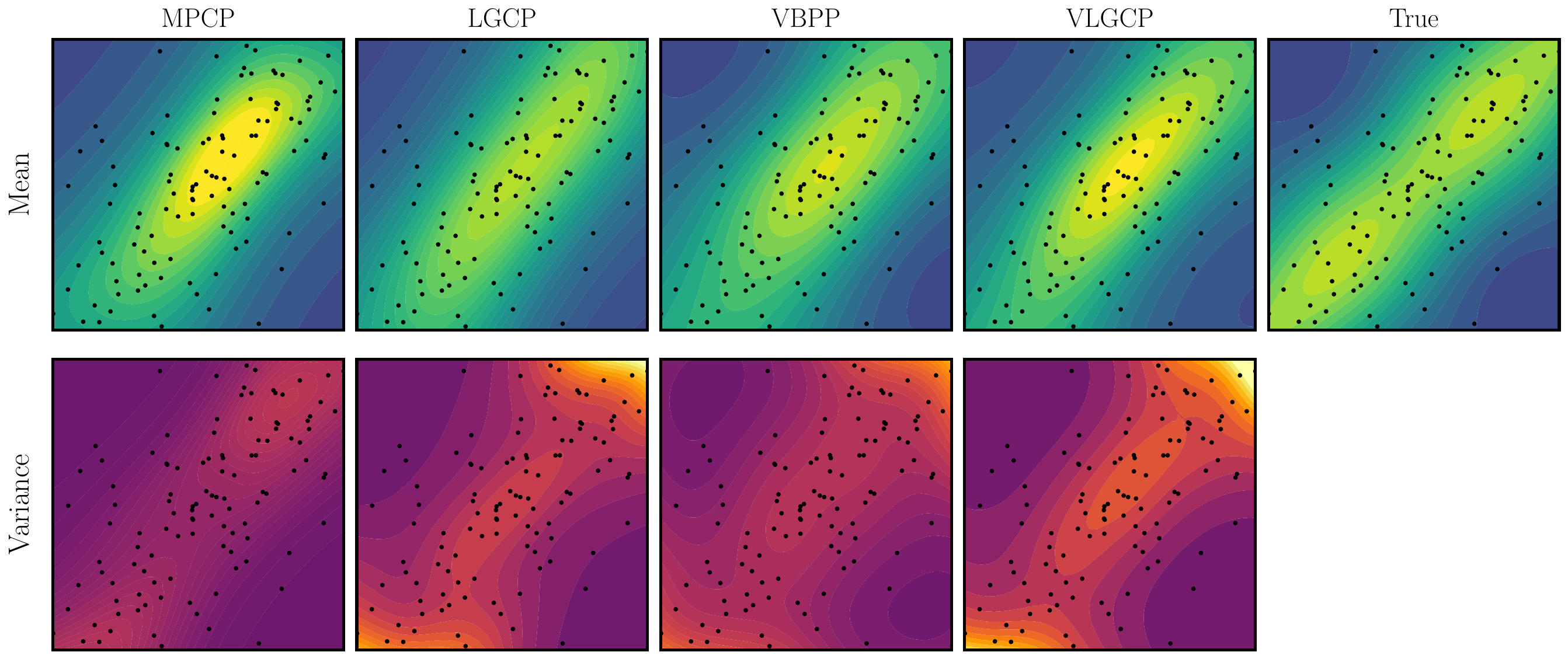}
  \caption{Posterior mean (top) and variance (bottom) inferred from the observed events (black dots), with the true intensity function $\lambda_2$ shown in the top-right panel. Lighter colors indicate higher mean and variance.}
  \label{fig:2D_comparison}
\end{figure}

The posterior intensities obtained for $\lambda_1$ are shown in \Cref{fig:1D_comparison}. For $\lambda_2$, we show the posterior mean and variance of the intensity in \Cref{fig:2D_comparison}. The posterior means are comparable across methods, with MPCP yielding slightly smaller posterior uncertainty. 

\begin{figure}[h]
  \centering
  \includegraphics[width=0.7\textwidth]{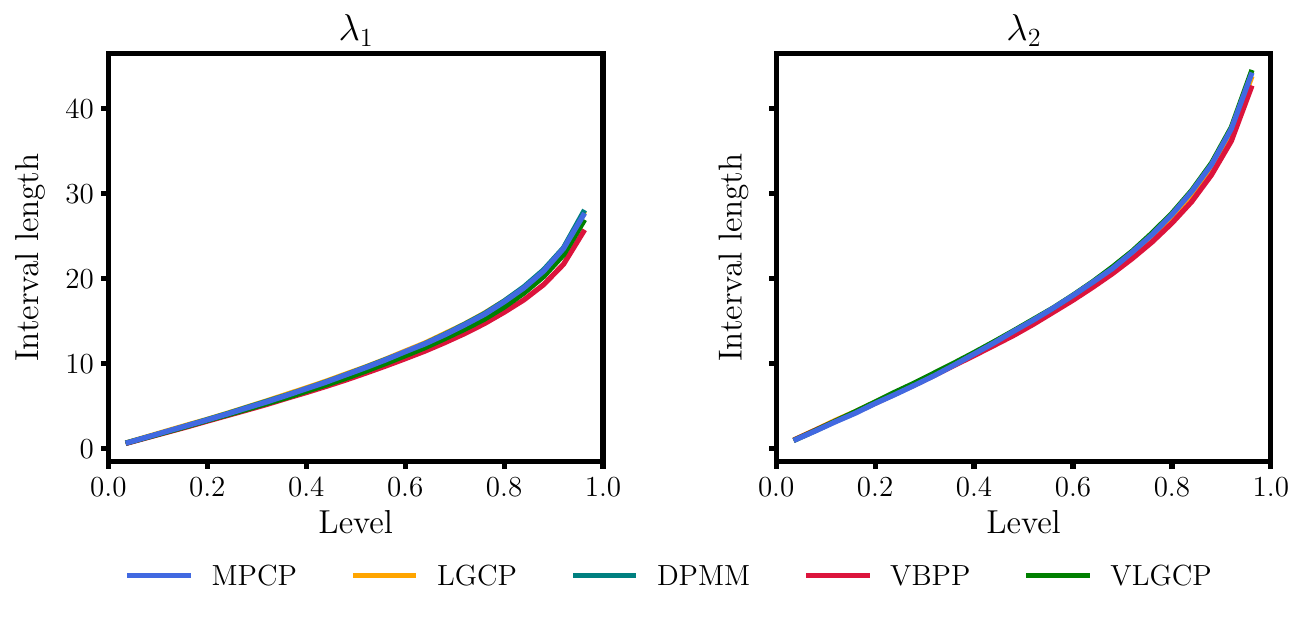}
  \caption{Lengths of the posterior credible intervals for $\gamma = \int_\mathcal{X} \lambda(x)\textup{d}x$ corresponding to level $q$.}
  \label{fig:length_plots}
\end{figure}

For quantitative comparison, we consider several performance measures. First, we use the integrated squared error,
\begin{align*}
    L_2 = \frac{1}{|\mathcal{X}|} \int_\mathcal{X} (\lambda^*(x) - \overline{\lambda}(x))^2\, \textup{d}x,
\end{align*}
where $\lambda^*$ denotes the true intensity and
$\overline{\lambda}(x) := \mathbb{E}[\lambda(x)\mid x_{1:n}]$
denotes the posterior mean intensity. We additionally consider the joint test log-likelihood,
\begin{align*}
    L_\textup{test} = \log p(x_{1:n_\textup{test}}^{\textup{test}} \mid x_{1:n}) = \log \mathbb{E}_{\lambda\mid x_{1:n}} \left[ \exp\left\{ -\int_\mathcal{X} {\lambda}(x) \textup{d}x \right\} \prod_{i=1}^{n_\textup{test}} {\lambda}(x_i^{\textup{test}}) \right],
\end{align*}
where $x_{1:n_\textup{test}}^\textup{test} = \{x_i^\textup{test}\}_{i=1}^{n_\textup{test}}$ denotes the set of test points. Finally, we assess the calibration of the posterior distribution of the average count ${\gamma} = \int_\mathcal{X}{\lambda}(x)\textup{d}x$ by considering the empirical coverage of the true mean count $\gamma^* := \int_\mathcal{X}\lambda^*(x)\textup{d}x$. Specifically, the coverage measures the proportion of simulated datasets for which $\gamma^*$ falls within the posterior $q$-level credible interval for ${\gamma}$. The resulting coverage values are reported in \Cref{fig:coverage_main}. We also report the corresponding credible-interval lengths in \Cref{fig:length_plots}. We find that the interval lengths obtained with MPCP are comparable to those of the competing methods.

\subsection{Real-Data Experiments}

\begin{figure}[h]
    \centering
    \includegraphics[width=\textwidth]{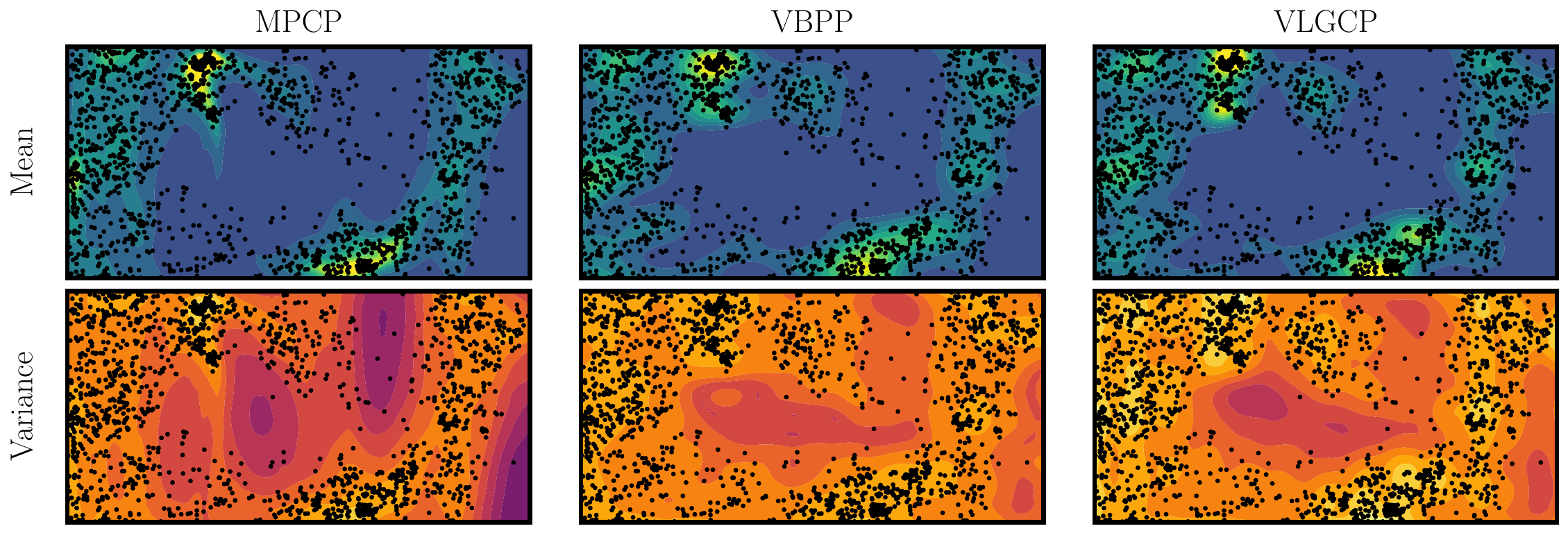}
    \caption{Posterior mean (top) and variance (bottom) of MPCP, VBPP and VLGCP (left to right) on the \textsc{bei} dataset. Lighter colors indicate higher mean and variance.}
    \label{fig:bei_predictions}
\end{figure}

We consider the publicly available \textsc{bei} dataset, which records the locations of trees in the tropical rainforest on Barro Colorado Island \citep{hubbell1983diversity}. We also consider the publicly available \textsc{earthquake} dataset from the U.S. Geological Survey Earthquake Catalog~\citep{usgs_earthquake_catalog}, which records earthquakes with Richter magnitude greater than 4 over the 10-year period from 2000 to 2009. For both datasets, we use the setup described in \Cref{sec:methodology}, with the prior $\pi(\gamma) \propto {\gamma^{-1}}$ for MPCP. For VBPP and VLGCP, we use $20^2 = 400$ inducing points/bins for the \textsc{bei} dataset and $10^3 = 1{,}000$ inducing points/bins for the \textsc{earthquake} dataset. For the latter, increasing the number of inducing points or bins led to memory issues, as it requires inverting a matrix exceeding $1{,}000 \times 1{,}000$ in size. The posterior mean and variance of the intensity for the \textsc{bei} dataset are shown in \Cref{fig:bei_predictions}. We find that the posterior mean and variance are comparable across all methods.

\begin{figure}[t!]
  \centering
  \includegraphics[width=\textwidth]{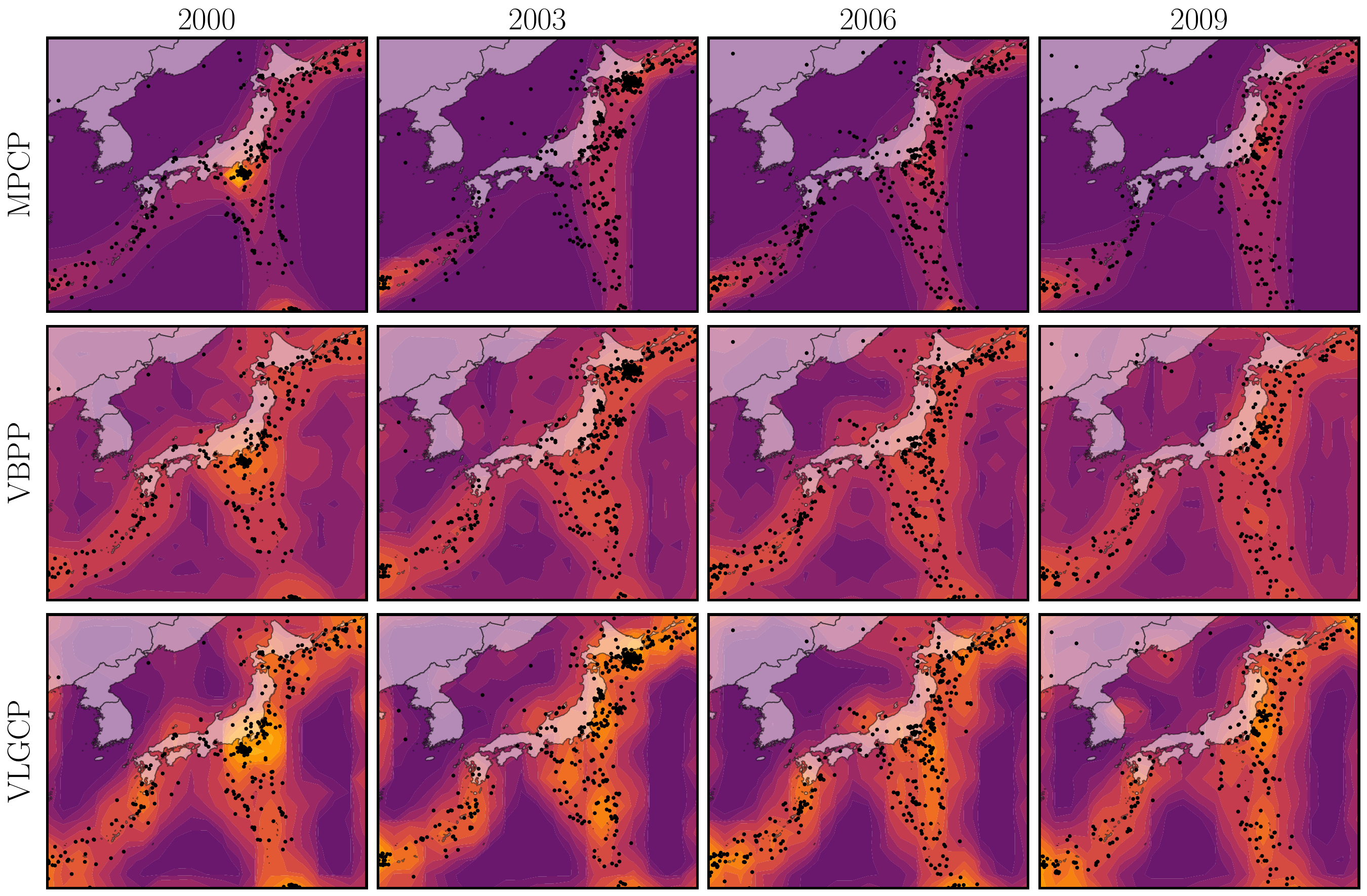}
  \caption{Posterior variance of the intensity over time for \textsc{earthquake} data for MPCP, VBPP, and VLGCP (top to bottom) at four time points (left to right), with black points indicating the observed earthquake events. Lighter colors indicate larger variance.}
  \label{fig:eq_var}
\end{figure}

For the \textsc{earthquake} dataset, we show the posterior mean and variance over time in \Cref{fig:eq_mean} and \Cref{fig:eq_var}, respectively. For each year, we evaluate the intensity at the midpoint of the year; for example, the intensity for 2000 is evaluated on 1 July 2000. As for the \textsc{bei} dataset, the posterior mean is comparable across all methods. We find that the uncertainty for GP approaches is slightly larger in this setting, likely due to the variational approximations.

We quantitatively analyze the performance of the methods by splitting the datasets into two sets of equal size, resulting in $n=1802$ training points for the \textsc{bei} dataset and $n=4477$ training points for the \textsc{earthquake} dataset, following \cite{aglietti2019structured, sellier2023sparse}. We then evaluate the test log-likelihood on the test dataset, as defined in Appendix~\ref{app:synthetic_experiments}.

\subsection{Sparse Approximations of VBPP and VLGCP} \label{app:vb_ablation}

\begin{figure}[h!]
  \centering
  \includegraphics[width=0.8\textwidth]{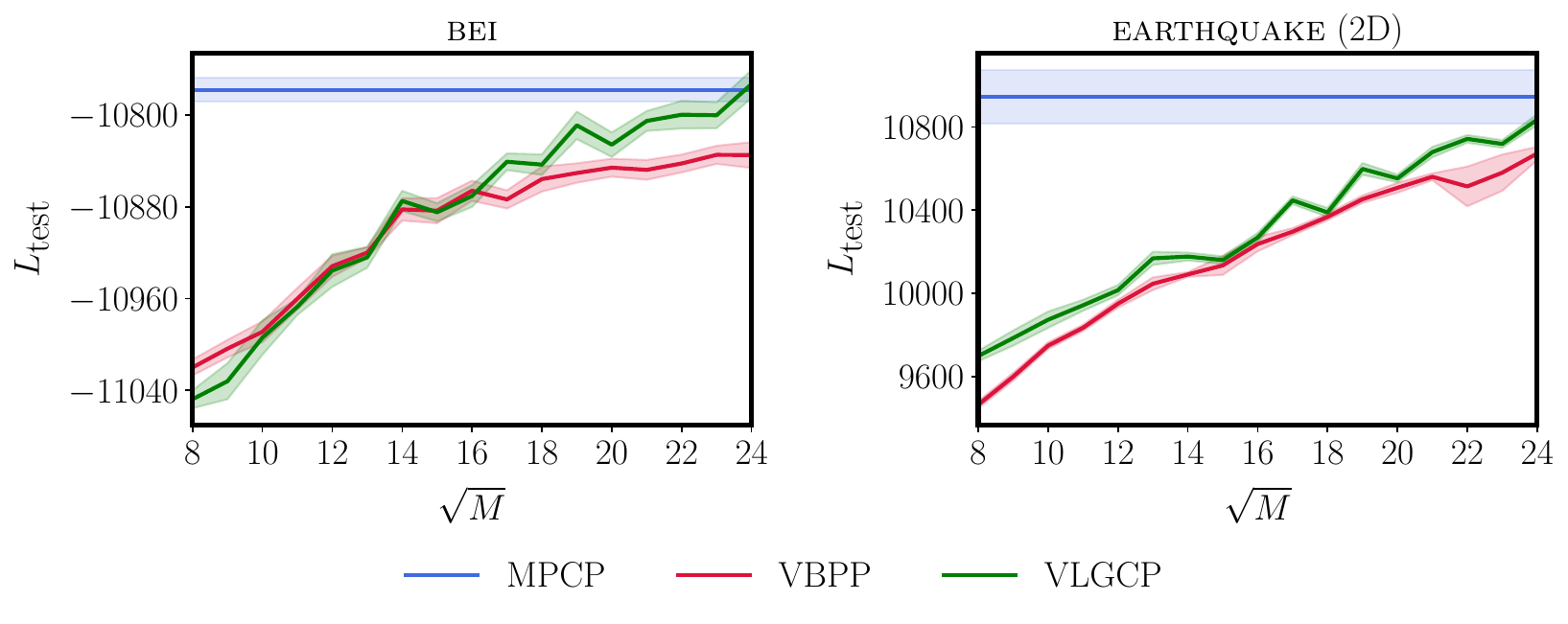}
  \caption{Mean and standard error of $L_\textup{test}$ for VBPP and VLGCP as the number of inducing points/bins increases, compared with MPCP, for the \textsc{bei} (left) and two-dimensional \textsc{earthquake} (right) datasets.}
  \label{fig:vb_ablation}
\end{figure}

Since VBPP and VLGCP rely on sparse approximations based on inducing points and bins, respectively, we investigate how the number of inducing points/bins affects intensity-estimation performance. We consider the \textsc{bei} dataset and the two-dimensional version of the \textsc{earthquake} dataset, such that both datasets have $d=2$ covariates. \Cref{fig:vb_ablation} shows that performance improves as the square root of the number of inducing points/bins, $\sqrt{M}$, increases. 
This also helps explain the performance gap observed for the three-dimensional \textsc{earthquake} dataset in \Cref{sec:realdata}. For the considered range of inducing points/bins, increasing their number already results in substantially higher run-times than MPCP (see \Cref{tab:real_data}). Nevertheless, MPCP, which does not rely on sparse approximations, outperforms or is competitive with both methods in terms of $L_\textup{test}$. Note that Figure \ref{fig:vb_ablation} is plotting $\sqrt{M}$ whereas run-time increases as $M^3$, so we are in the region of diminishing returns.

\end{document}